\documentclass[review]{elsarticle}
\usepackage{graphicx}%
\usepackage{multirow}%
\usepackage{amsmath,amssymb,amsfonts}%
\usepackage{amsthm}%
\usepackage{mathrsfs}%
\usepackage[title]{appendix}%
\usepackage{xcolor}%
\usepackage{textcomp}%
\usepackage{manyfoot}%
\usepackage{booktabs}%
\usepackage{algorithm}%
\usepackage{algorithmicx}%
\usepackage{algpseudocode}%
\usepackage{listings}%

\usepackage{hyperref}
\usepackage{graphicx}
\usepackage{amsmath, amssymb}
\usepackage{booktabs}
\usepackage{makecell}   
\usepackage{threeparttable} 
\usepackage{siunitx}
\usepackage{adjustbox}  
\usepackage{amsmath} 
\usepackage{rotating}     
\usepackage{booktabs}
\usepackage{microtype}
\usepackage{geometry}
\usepackage{microtype}
\usepackage{inputenc}
\usepackage{geometry} 
\usepackage[T1]{fontenc}
\usepackage{lmodern}
\usepackage{tabularx}
\newcolumntype{Y}{>{\centering\arraybackslash}X}
\usepackage{caption}      
\usepackage{array}
\newcolumntype{C}{>{\centering\arraybackslash}p{1.2cm}}

\usepackage{pifont}   
\usepackage{array}

\hypersetup{
    colorlinks=true,
    linkcolor=black,
    citecolor=blue,
    urlcolor=blue,
    pdftitle={DiffKD-DCIS: Predicting Upgrade of Ductal Carcinoma In Situ},
    pdfauthor={Tao Li et al.},
    unicode=true
}

\usepackage{lineno,hyperref}
\modulolinenumbers[5]

\journal{Journal of \LaTeX\ Templates}

\begin{document}

\begin{frontmatter}

\title{DiffKD-DCIS: Predicting Upgrade of Ductal Carcinoma In Situ with Diffusion Augmentation and Knowledge Distillation}
\tnotetext[mytitlenote]{Fully documented templates are available in the elsarticle package on \href{http://www.ctan.org/tex-archive/macros/latex/contrib/elsarticle}{CTAN}.}

\author{Tao Li\textsuperscript{1\#}, Qing Li\textsuperscript{1\#}, Na Li\textsuperscript{2\#}, Hui Xie\textsuperscript{2,3*}}
\address{1. School of Medical Imaging, Laboratory Medicine and Rehabilitation, Xiangnan University, Chenzhou,423000, P. R. China\\ 2. Department of Radiotherapy, Affiliated Hospital (Clinical College) of Xiangnan University, Chenzhou,423000, P.R.China\\ 3. Faculty of Applied Sciences, Macao Polytechnic University, Macao,999078, P.R.China//}
\fntext[myfootnote]{Since 1880.}

\author[]{}
\ead[]{}

\cortext[mycorrespondingauthor]{Correspondingauthor: Hui Xie}
\ead{h.xie@xnu.edu.cn}

\address[mymainaddress]{1600 John F Kennedy Boulevard, Philadelphia}
\address[mysecondaryaddress]{360 Park Avenue South, New York}

\begin{abstract}
Objective: Accurately predicting the upgrade of ductal carcinoma in situ (DCIS) to invasive ductal carcinoma (IDC) preoperatively is crucial for surgical planning. However, traditional deep learning methods face significant challenges due to limited ultrasound data and poor generalization capability. This study proposes DiffKD-DCIS, a novel framework integrating conditional diffusion modeling with teacher–student knowledge distillation, to improve the accuracy and robustness of DCIS upgrade prediction.

Methods: Our approach employs a three-stage pipeline. First, a conditional diffusion model generates high-fidelity ultrasound images for data augmentation by incorporating multimodal conditions—including textual prompts, tumor region masks, and class labels. Second, a deep teacher network is trained on both original and synthesized data to extract robust features. Finally, a compact student network learns from the teacher via knowledge distillation, enhancing generalization while maintaining computational efficiency. The framework was evaluated on a multicenter dataset comprising 1,435 DCIS cases from three independent medical centers. Additionally, three radiologists with varying levels of experience independently interpreted the cases for human–machine comparative analysis.

Results:Synthesized images achieved a peak signal-to-noise ratio (PSNR) of $\SI{22.65 \pm 2.31}{dB}$ and structural similarity index (SSIM) of $0.87 \pm 0.04$. The student network contained only \SI{37.3}{\percent} of the teacher's parameters and exhibited a $2.7\times$ faster inference speed.  On two external test sets, DiffKD-DCIS achieved AUCs of 0.812 (95\% CI: 0.787--0.837) and 0.809 (95\% CI: 0.760--0.858), with accuracies of \SI{78.5}{\percent} (95\% CI: \SI{76.3}{\percent}--\SI{80.7}{\percent}) and \SI{78.0}{\percent} (95\% CI: \SI{73.4}{\percent}--\SI{82.6}{\percent}), respectively. Ablation studies demonstrated that diffusion augmentation without knowledge distillation achieved AUCs of 0.776 (95\% CI: 0.748--0.804) and 0.772 (95\% CI: 0.723--0.821), while knowledge distillation with traditional augmentation yielded AUCs of 0.742 (95\% CI: 0.714--0.770) and 0.738 (95\% CI: 0.689--0.787), underscoring the synergistic benefits of combining both components.

Human--machine comparison revealed that the model's accuracy was comparable to that of senior radiologists ($P = 0.215$) and significantly outperformed junior radiologists ($P = 0.045$).

Conclusion: DiffKD-DCIS effectively addresses the challenges of data scarcity and overfitting in DCIS upgrade prediction. The integration of conditional diffusion modeling and knowledge distillation establishes a robust framework for medical image analysis, achieving diagnostic performance on par with senior radiologists. This highlights its significant clinical potential and value for widespread adoption in practice.
\end{abstract}

\begin{keyword}
\texttt{Ductal carcinoma in situ (DCIS)\sep Ultrasound classification\sep Diffusion model\sep Knowledge distillation\sep Multimodal learning\sep Breast cancer}
\MSC[2010] 00-01\sep  99-00  

\(\#\) :These authors contributed equally to this work.

\end{keyword}

\end{frontmatter}

\section{Introduction}
\label{１}

Ductal carcinoma in situ (DCIS) is a non-invasive form of breast cancer confined within the basement membrane of the mammary ducts, wherein malignant cells proliferate without breaching the membrane. Consequently, treatment strategies are relatively conservative, typically involving breast-conserving surgery combined with radiotherapy. In contrast, once DCIS upgrades to invasive ductal carcinoma (IDC), axillary lymph node evaluation—such as sentinel lymph node biopsy or axillary dissection—becomes necessary, highlighting substantial differences in treatment planning and prognosis\cite{solin2019management}  \cite{shah2016management}. However, core needle biopsy (CNB), the current gold standard for preoperative diagnosis, has limited sampling coverage and fails to detect occult invasive components in approximately 15\(\%\)–30\(\%\) of cases, leading to postoperative pathological upgrades\cite{britton2009one} \cite{slostad2022concordance} \cite{yates2015subclonal}\cite{rahbar2022imaging} \cite{yoon2024ai}. This diagnostic–pathological discrepancy results in suboptimal surgical decisions: some patients experience delayed axillary intervention due to underestimation of invasion risk, while others undergo unnecessary axillary procedures due to overestimation. Hence, developing a reliable tool for preoperative prediction of DCIS upgrade is of critical clinical importance for optimizing surgical decision-making.

Existing studies have extensively explored DCIS upgrade prediction models based on clinicopathological factors (e.g., patient age, maximum tumor diameter, nuclear grade), yet their predictive performance remains generally limited (AUC typically 0.70–0.75)\cite{brennan2011ductal}\cite{jakub2017validated} \cite{park2013nomogram}. Other studies have attempted to integrate imaging features (e.g., calcification morphology on mammography, enhancement patterns on MRI) with clinical data to build hybrid models. However, due to the reliance on manual delineation of imaging features, these approaches suffer from subjectivity and low efficiency, achieving only modest improvements with AUCs ranging from 0.73 to 0.78—still falling short of the clinically desirable threshold (>0.8)\cite{schulz2013prediction} \cite{alaeikhanehshir2024application} \cite{mayfield2024presurgical}. Ultrasound, a radiation-free, low-cost, and widely accessible imaging modality, holds particular advantages for detecting DCIS in dense breasts and non-calcified lesions\cite{su2017non}. Nevertheless, most existing deep learning models based on ultrasound directly train on limited retrospective data, resulting in frequent overfitting and poor generalization due to small sample sizes and heterogeneous image quality\cite{durur2023artificial}. Some works have employed conventional data augmentation (e.g., rotation, flipping) to expand ultrasound datasets, yet these methods often fail to preserve critical microscopic structures (e.g., microcalcifications, ductal dilation), inadvertently introducing noise and reducing model specificity \cite{medghalchi2025synthetic} \cite{garcea2023data}. Although conditional diffusion models have demonstrated superior detail-preserving capabilities in medical image synthesis\cite{singh2023high}, their application to ultrasound image augmentation for DCIS remains unexplored. Similarly, while teacher–student networks leveraging knowledge distillation have shown promise in improving small-sample learning in other domains\cite{hinton2015distilling} \cite{tang2020understanding} \cite{zhang2025few}, their use in DCIS upgrade prediction is still a research gap.

To address the aforementioned challenges, this study proposes a deep learning framework based on conditional diffusion models and teacher–student networks. First, a conditional diffusion model is employed to generate and enhance original ultrasound images with high fidelity, thereby expanding the sample size while preserving critical microscopic lesion details—such as irregular margins and heterogeneous internal echogenicity. Subsequently, both the generated and original images are jointly fed into a teacher–student network for training: the teacher network, equipped with a deep convolutional architecture, extracts high-level lesion features, whereas the student network learns generalizable representations via knowledge distillation, thereby mitigating overfitting. By integrating multimodal conditional information—including textual prompts, tumor region masks, and class labels—an end-to-end predictive model is constructed, ultimately enabling precise preoperative prediction of upgrade status in patients diagnosed with DCIS via CNB. This study aims to enhance ultrasound data quality through conditional diffusion modeling and leverage the feature-learning capability of teacher–student networks to improve the performance of DCIS upgrade prediction, thereby providing a reliable basis for clinical precision decision-making.

Our main contributions are threefold: 

1. We develop a conditional diffusion model that integrates textual prompts, tumor masks, and class labels to generate high-fidelity ultrasound images—effectively mitigating data scarcity while preserving clinically relevant features.  

2. We propose a two-stage training strategy: a teacher network learns robust features from the augmented data, and a compact student network achieves enhanced generalization through knowledge distillation.  

3. We establish a comprehensive evaluation protocol, assessing both image generation quality and clinical classification performance on a multi-center dataset, demonstrating significant improvements over existing methods.

\section{Related Work}
\label{}

\subsection{DCIS Upgrade Prediction}
\label{subsec1}

Traditional DCIS upgrade prediction methods primarily rely on clinicopathological factors. Several nomograms have been developed that integrate features such as patient age, tumor size, nuclear grade, and comedonecrosis\cite{brennan2011ductal} \cite{jakub2017validated} \cite{park2013nomogram}. These models are typically constructed using logistic regression or decision trees to quantify upgrade risk. However, their performance remains suboptimal for clinical implementation, with AUC values generally ranging between 0.70 and 0.75—largely limited by the restricted feature set and the omission of imaging information.

Recently, image-based predictive models have garnered increasing attention. Studies have explored mammographic features\cite{yoon2024ai} \cite{hou2019prediction} (e.g., calcification morphology and distribution), MRI radiomics\cite{kim2025mri} \cite{yao2024kinetic} \cite{lee2022prediction}(e.g., texture heterogeneity and kinetic enhancement patterns), and ultrasound characteristics \cite{zhu2024ultrasound} (e.g., lesion margin, echogenicity pattern, and vascularity). For instance, Hou et al. \cite{hou2019prediction} built an upgrade prediction model using clustered calcification features from mammograms, achieving a precision of only 72\(\%\); however, the approach heavily depended on radiologists’ subjective interpretation, resulting in high inter-observer variability (Kappa < 0.6). Similarly, Lee et al.\cite{lee2022prediction} extracted over 100 radiomic features from MRI, but the model exhibited degraded generalizability on external validation—AUC dropped from 0.82 (internal) to 0.71 (external)—highlighting the challenge posed by multicenter data heterogeneity.

Deep learning methods have shown promise in automatically extracting discriminative features from medical images. Qian et al.\cite{qian2021application}  developed a CNN-based classification model for DCIS using ultrasound images, with ResNet-50 as the backbone, achieving an AUC of 0.85 on a single-center dataset. Yet, generalization to external datasets was limited (AUC dropped to 0.68), primarily due to overfitting caused by insufficient training data. Similar issues persist across other studies—for example, Toa et al. \cite{toa2024deep} attempted to enhance CNNs with attention mechanisms but still faced bottlenecks in small-sample learning. A key limitation of existing deep learning approaches is their reliance on scarce annotated data, leading to overfitting and reduced clinical applicability. Moreover, data imbalance—where upgrade cases are typically far fewer than non-upgrade cases—further exacerbates model bias, resulting in high false-positive rates\cite{hashiba2023prediction}.

\subsection{Applications of Diffusion Models in Medical Imaging}
\label{subsec2}

Diffusion models~\cite{shen2024imagpose,shen2025imagdressing,shenlong,shen2025imaggarment} have emerged as powerful generative frameworks capable of producing high-fidelity images, with Denoising Diffusion Probabilistic Models (DDPMs) being a representative approach\cite{ho2020denoising}. In medical imaging, diffusion models have been applied to diverse tasks, including image reconstruction \cite{song2020score}, segmentation \cite{wolleb2022diffusion}\cite{usman2024brain} , and data augmentation\cite{wu2024diffusion}. For instance, Wolleb et al. {wolleb2022diffusion} employed diffusion models to reconstruct high-resolution images from low-dose CT scans, achieving higher PSNR (28.4 dB vs. 25.1 dB) than GANs while preserving lesion details such as lung nodule margins. Stacke et al.\cite{usman2024brain} integrated diffusion models into a segmentation framework for brain tumor delineation, reporting a 15.9\(\%\) improvement in SSIM over the baseline—demonstrating their robustness against MRI noise and anatomical variability. Nevertheless, most existing applications focus on CT and MRI modalities, with limited exploration in ultrasound imaging—primarily due to the inherent speckle noise and complex texture patterns in ultrasound, which lead to training instability and insufficient generation diversity\cite{dominguez2024diffusion}.

Conditional diffusion models extend this capability by incorporating auxiliary information to guide the generation process. For example, Qin et al. \cite{qin2025btsegdiff} utilized segmentation masks as conditions for brain MRI synthesis, improving anatomical consistency by 19.2\(\%\) in SSIM and boosting downstream brain tumor segmentation accuracy by 5.3\(\%\). Masutani et al.\cite{bluethgen2025vision}  leveraged text prompts for chest X-ray synthesis, integrating semantic information through CLIP embeddings to enable controllable diversity in generation (e.g., producing semantically consistent lesion variants via prompt manipulation). Our work advances this direction by integrating multiple condition types—specifically tailored for DCIS characterization in ultrasound—including textual prompts, tumor masks, and class labels. In contrast to prior efforts, our model employs custom prompt templates (e.g., “ultrasound image of carcinoma in situ with stable features”) and a mask encoder to explicitly preserve tumor-region characteristics—a first-of-its-kind attempt in ultrasound imaging. Moreover, our diffusion process is configured with 1,000 sampling steps and a cosine noise schedule to optimize training stability\cite{nichol2021improved}.

\subsection{Applications of Knowledge Distillation in Medical AI}
\label{subsec3}

Knowledge distillation (KD) enables a compact student network to learn from a larger teacher network, improving computational efficiency without significant performance degradation\cite{hinton2015distilling}. In medical imaging, KD has been applied to various tasks, including disease classification\cite{zheng2023kd_convnext}, lesion detection\cite{jin2020ra}, and segmentation\cite{gorade2024rethinking} . For instance, Bai et al. \cite{zheng2023kd_convnext} applied KD to pulmonary nodule classification, transferring soft labels from a Swin Transformer teacher to a ConvNeXt student—achieving a 50\(\%\) reduction in parameters with only a 1.8\(\%\) drop in accuracy. Tang et al.  \cite{jin2020ra} extended KD to 3D imaging by incorporating attention distillation to emphasize diagnostically critical regions, improving recall by 10.2\(\%\) in liver tumor detection. Additionally, Liu et al.\cite{gorade2024rethinking} employed intermediate-layer distillation for kidney and liver tumor segmentation, maintaining Dice scores within 1.5\(\%\) of the teacher despite substantial parameter compression—demonstrating KD’s robustness on 3D volumetric data. However, most existing approaches rely on conventional augmentation techniques (e.g., flipping, cropping) and do not leverage advanced generative models for data expansion, limiting their potential in data-scarce scenarios\cite{kebaili2023deep} —particularly in privacy-sensitive healthcare settings, where synthetic data have been shown to improve generalization by up to 15\(\%\)\cite{qin2025btsegdiff}.

Our work uniquely integrates conditional diffusion-based data augmentation with knowledge distillation, establishing a synergistic framework that jointly addresses data scarcity and model efficiency in medical image analysis. Unlike classical distillation (e.g., Hinton et al.\cite{hinton2015distilling}), we adopt a softened label distribution with a temperature scaling factor of 3.0 and a hybrid loss function weighted by \(\alpha\) = 0.7, specifically tuned to accommodate the label noise and variability inherent in clinical data [45]. Furthermore, our teacher network employs a 4-layer convolutional architecture (with channels progressively increasing to 512), while the student network is streamlined to 3 layers (with maximum channels reduced to 128)—aligning with real-world clinical deployment constraints in terms of memory and latency\cite{ghojogh2024neural}. To further reduce training overhead, we apply an additional fine-tuning phase on the distilled student, optimizing convergence efficiency without compromising performance \cite{hinton2015distilling}.

\section{Materials and Methods}

This study has been approved by the Ethics Committee of Xiangnan University (IRB No.: 2023YX11014). Given the retrospective and anonymized nature of the data, the committee granted a waiver of informed consent. All data handling procedures strictly adhere to the principles of the Declaration of Helsinki and the Ethical Review Measures for Biomedical Research Involving Human Subjects (China). Data anonymization included: (1) removal of all direct personal identifiers (e.g., name, ID number, telephone number); (2) temporal obfuscation of indirect identifiers (e.g., examination date and birth date via fixed-offset shifting); and (3) de-identification of device serial numbers and institutional information in image metadata through encoding.

\subsection{Data Acquisition and Quality Control}

We retrospectively collected ductal carcinoma in situ (DCIS) cases from three medical institutions—Affiliated Hospital of Xiangnan University, Affiliated Rehabilitation Hospital of Xiangnan University, and the Third People’s Hospital of Chenzhou—between January 2015 and January 2025. All cases underwent preoperative breast ultrasound examination, and postoperative histopathological findings served as the reference standard.  

Inclusion criteria were: (1) ultrasound images of sufficient diagnostic quality; and (2) complete clinical and pathological records. Exclusion criteria included: (1) poor image quality or severe artifacts; (2) prior history of malignant tumor treatment in the ipsilateral breast; and (3) missing clinical or pathological data.  

A total of 1,435 pathology-confirmed DCIS cases were ultimately enrolled. All ultrasound images were acquired using a standardized protocol across multiple vendor platforms, including Philips, Siemens, Canon Medical, GE Healthcare, Hitachi, Mindray, and Esaote. Lesion annotations were performed by radiologists with >10 years of experience in breast imaging interpretation, and the ground truth for upgrade status (i.e., presence of microinvasion or invasive carcinoma on surgical pathology) was determined by final histopathological diagnosis.

To augment the training dataset (804 real images: 438 upgraded, 366 non-upgraded), we generated synthetic ultrasound images using the conditional diffusion model with a differential strategy: 8 per non-upgraded case (total 2,928) and 5 per upgraded case (total 2,190), resulting in 5,118 synthetics and an approximate real-to-synthetic ratio of 1:6.36. This ratio was selected to address class imbalance original upgraded:non-upgraded $\approx$1.2:1), achieving a post-augmentation upgrade rate of $\sim$45\%, while preventing overfitting to synthetic artifacts. For detailed ablation experiments validating this choice, including tests on alternative ratios and multi-metric evaluations, see Supplementary Material S1.

\subsection{DiffKD-DCIS Framework}

To address the dual challenges of limited sample size and image quality heterogeneity in DCIS upgrade prediction, this paper proposes DiffKD-DCIS (Diffusion-based Knowledge Distillation for DCIS Upgrade Prediction), a novel framework. As illustrated in Figure 1, the framework consists of three synergistic modules:  

(1) Conditional Diffusion Generation Module: A high-fidelity ultrasound image generation model conditioned on clinical ultrasound images and their corresponding pathological labels, specifically designed to preserve DCIS-relevant diagnostic features—such as clustered microcalcifications, ductal dilatation, and subtle margin microstructures;  

(2) Teacher Network: A high-capacity deep neural network trained on a hybrid dataset combining original and diffusion-augmented images, serving as the knowledge provider;  

(3)Student Network: A lightweight network that transfers discriminative feature representations from the teacher network via anatomy-aware knowledge distillation, thereby enhancing generalization and robustness under limited annotated data. The end-to-end pipeline establishes a closed-loop “Generate-Augment-Distill” framework, jointly ensuring the fidelity of data augmentation and the efficiency of model learning. For ablation studies, we evaluated isolated components: (1) Diffusion augmentation without KD: Directly training the student network on the augmented dataset (real + synthetic images) without distillation from the teacher. (2) KD with traditional augmentation: Applying traditional data augmentation techniques (e.g., rotation, flipping, brightness jitter) to the original dataset, training the teacher network, and then performing knowledge distillation to the student network. These variants were compared against the full framework to quantify individual contributions.

\begin{figure}
    \centering
    \includegraphics[width=1\linewidth]{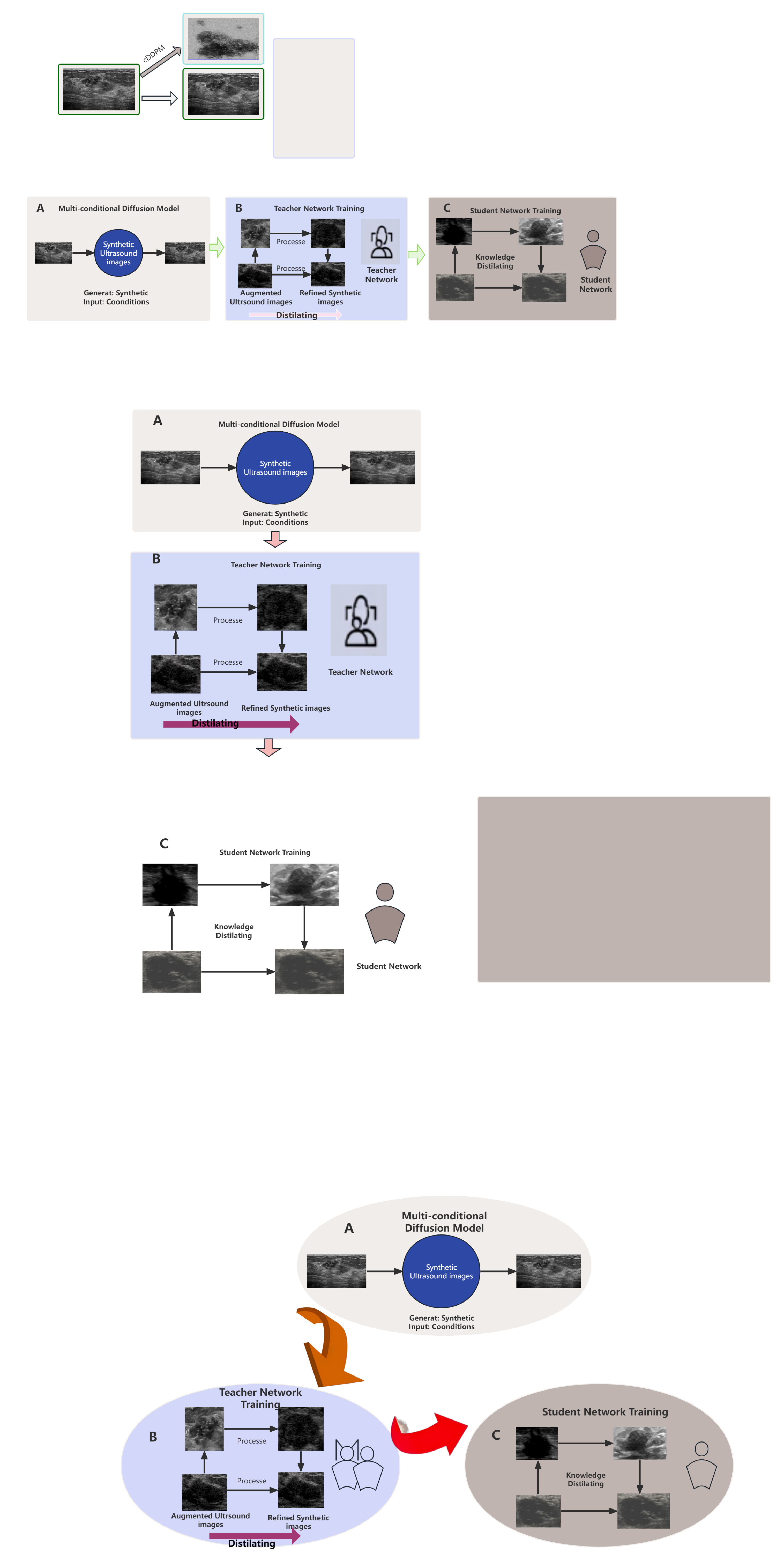}
    \caption{Overall architecture of DiffKD-DCIS. (A) Multi-conditional diffusion model guided by text prompts, tumor masks, and upgrade labels to generate high-fidelity synthetic ultrasound images. (B) Teacher network trained on real + synthetic augmented dataset. (C) Lightweight student network distilled from the teacher, maintaining comparable performance while outperforming all baselines on two independent external test sets.}
    \label{fig:placeholder}
\end{figure}

\subsubsection{Conditional Diffusion Models}

This study employs a Latent Diffusion Model (LDM) as the core architecture for image generation, with its key innovation lying in shifting the diffusion process from pixel space to a low-dimensional latent space learned by an ultrasound-optimized Variational Autoencoder (US-VAE), thereby improving both generation efficiency and image fidelity, while multimodal conditional guidance ensures that the synthesized ultrasound images exhibit high clinical relevance and structural fidelity; as illustrated in Figure 2, the overall architecture comprises Module A as the conditional diffusion backbone, Modules B, C, and D (within the U-Net) corresponding to residual blocks, middle blocks, and MLP components respectively, Module E as the US-VAE encoder, and Module D also serving as the US-VAE decoder—though this reuse of label “D” for two distinct modules may introduce ambiguity and is recommended for clarification or relabeling in the figure.

\begin{figure}
    \centering
    \includegraphics[width=1\linewidth]{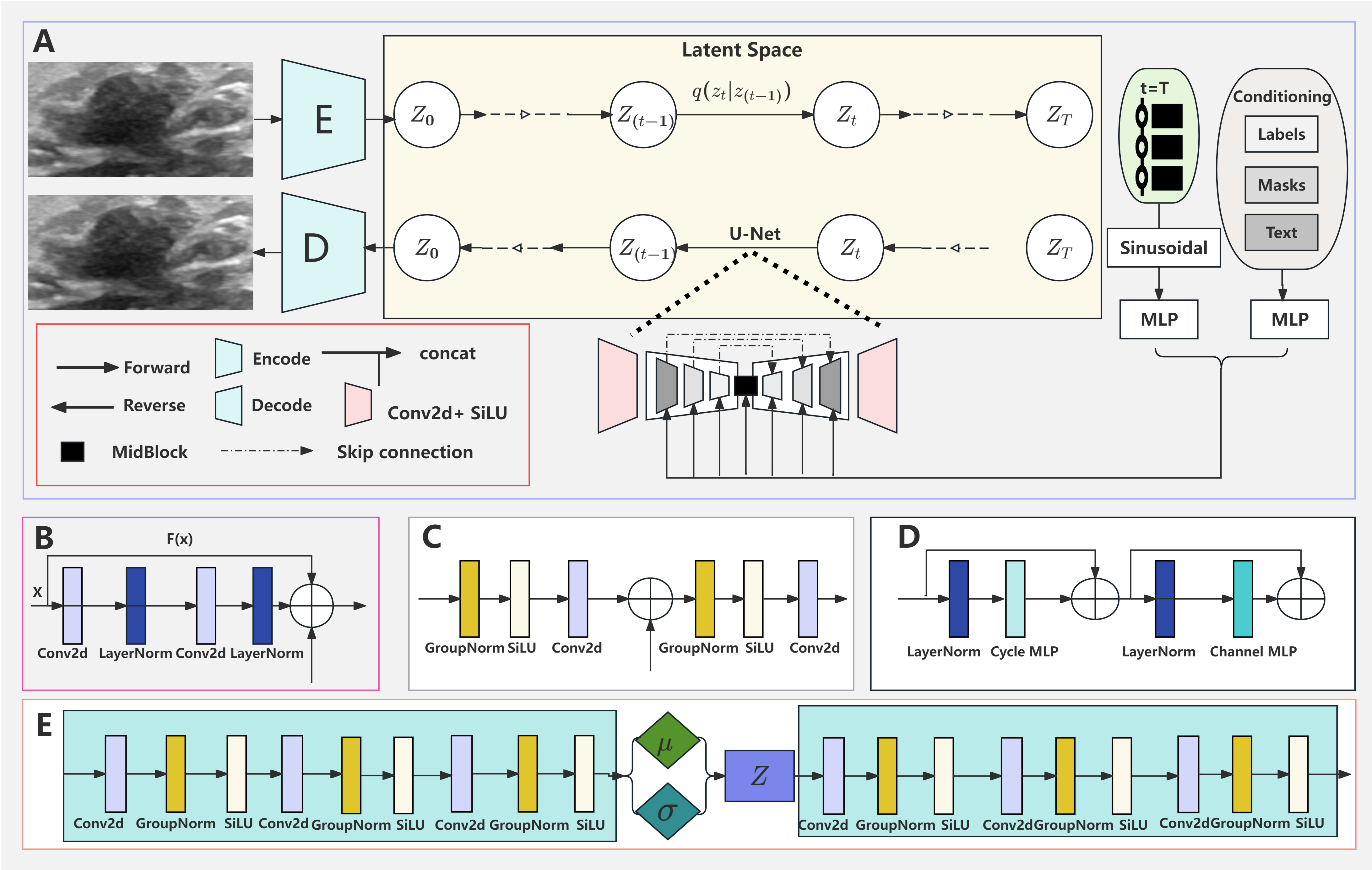}
     \caption{ Architecture of the Conditional Latent Diffusion Model for Ultrasound Image Synthesis. A: Conditional Diffusion Backbone — iteratively denoises the latent representation \(Z_t\) using the current time step t, noise level, and multi-modal conditioning signals (upgrade labels, tumor masks, and text prompts). Conditioning signals are injected via contact mechanisms in each U-Net block to precisely guide the generation of clinically relevant structures.  B: Residual Block of U-Net — composed of Conv2d, LayerNorm, and skip connections.  C: MidBlock of U-Net — employs GroupNorm, and SiLU activation to effectively model complex spatial and conditional relationships in the latent space. D: Multi-Layer Perceptron (MLP) — processes time-step embeddings and conditioning embeddings (labels \(\&\) text) before feeding them into the U-Net.  E: Ultrasound-Optimized Variational Autoencoder (VAE). Left: Encoder compresses raw ultrasound images into compact low-dimensional latent representations \(Z_0\); Right: Decoder reconstructs high-fidelity ultrasound images from the final denoised latent codes \(Z_0\).}
    \label{fig:placeholder}
\end{figure}

In the Conditional Latent Diffusion Model, we adopt a Denoising Diffusion Probabilistic Model (DDPM)\cite{ho2020denoising} with $T = 1000$ steps as the generative backbone. The forward process defines a Markov chain that gradually adds Gaussian noise to the original real image $x_0 \in \mathbb{R}^{H \times W}$ over $T$ steps, ultimately transforming it into pure noise $x_T$.

The transition distribution is given by:

\begin{equation}
q(x_t | x_{t-1}) = \mathcal{N}(x_t; \sqrt{1 - \beta_t} x_{t-1}, \beta_t I),
\end{equation}

Here, $\beta_t$ is the predefined noise schedule coefficient, following a cosine annealing scheme from $\beta_1 = 10^{-4}$ to $\beta_T = 0.02$:

\begin{equation}
\beta_t = \frac{\sin^2\!\left( \frac{\pi t}{T+1} \right)}{\sin^2\!\left( \frac{\pi (t+1)}{T+1} \right)},
\end{equation}

This scheduling scheme effectively balances the noise addition rate in early and late stages, improving training stability. The closed-form expression of $x_t$ at any time step $t$ can be written as:
\begin{equation}
x_t = \sqrt{\bar{\alpha}_t} x_0 + \sqrt{1 - \bar{\alpha}_t} \varepsilon,
\end{equation}
where $\bar{\alpha}_t = \prod_{i=1}^{t} (1 - \beta_i)$, and $\varepsilon \sim \mathcal{N}(0, I)$.

The goal of the reverse denoising process is to learn a neural network $\theta$ that can predict the noise $\varepsilon$ added during the forward process, given any noisy state $x_t$. Specifically, the reverse process is defined as:
\begin{equation}
p_\theta(x_{t-1} | x_t) = \mathcal{N}(x_{t-1}; \mu_\theta(x_t, t), \Sigma_\theta(x_t, t)),
\end{equation}
where the mean $\mu_\theta$ and variance $\Sigma_\theta$ are parameterized by the conditional U-Net network. We directly train the network $\varepsilon_\theta$ to predict the noise $\varepsilon$, optimizing by minimizing the following loss function:
\begin{equation}
L_{\text{diff}} = \mathbb{E}_{x_0, \varepsilon, t} \left\| \varepsilon - \varepsilon_\theta(x_t, t) \right\|^2.
\end{equation}

\subsubsection{Ultrasound-Optimized Variational Autoencoder (US-VAE)}

To transfer the diffusion process into latent space, we first introduce an ultrasound-optimized variational autoencoder (US-VAE). This module consists of an encoder $E$ and a decoder $D$, whose core function is to map a high-dimensional ultrasound image $x$ into a compact and semantically rich latent representation $z \in \mathbb{R}^{32 \times 32 \times 16}$, and during the generation phase, decode $z$ back into pixel space to obtain the synthesized image $x'$.

The architecture of encoder $E$ is shown in Figure~2(E), left part: it comprises four downsampling convolutional layers, each followed by GroupNorm and SiLU activation, finally outputting the mean $\mu$ and log-variance $\log\sigma^2$ of the latent variable:
\begin{equation}
[\mu, \log\sigma^2] = E(x),
\end{equation}

The latent variable $z$ is obtained via the reparameterization trick:
\begin{equation}
z = \mu + \sigma \odot \varepsilon, \quad \varepsilon \sim \mathcal{N}(0, I),
\end{equation}
where $\mu$ and $\sigma$ are the mean and standard deviation vectors output by encoder $E(x)$, and $\odot$ denotes element-wise multiplication.

The decoder $D$ has a symmetric structure to the encoder, comprising four upsampling convolutional layers, and finally outputs the reconstructed image $\hat{x}$:
\begin{equation}
\hat{x} = D(z),
\end{equation}

The training objective of US-VAE is to maximize the evidence lower bound (ELBO) of the data likelihood, combined with a perceptual loss to ensure that the latent representation effectively captures microstructural features of lesions:
\begin{equation}
\mathcal{L}_{\text{vae}} = 
\underbrace{\mathbb{E}_{q(z|x)}[\log p(x|z)]}_{\text{Reconstruction Loss}} 
- \underbrace{\beta \cdot KL(q(z|x) \| p(z))}_{\text{KL Divergence}} 
+ \underbrace{\lambda_{\text{perc}} \cdot \mathcal{L}_{\text{perceptual}}(x, \hat{x})}_{\text{Perceptual Loss}},
\end{equation}
where $\mathcal{L}_{\text{perceptual}}$ computes the perceptual similarity using intermediate features extracted from a pretrained teacher network, thereby constraining the generated images to preserve clinically relevant texture.

\subsubsection{Multimodal Conditional Integration and U-Net Architecture}

When performing diffusion in latent space, we design a conditional U-Net network $\varepsilon_\theta(z_t, t, c)$ that takes as input the noisy latent variable $z_t$, time step $t$, and a 384-dimensional multimodal conditional embedding $c$, obtained by concatenating outputs from three dedicated encoders—a 128-D class encoder for the binary upgrade label $y$, a 128-D mask encoder for the tumor region mask $m$, and a 128-D text encoder that compresses a 512-D CLIP-based embedding of clinical prompts (e.g., ``irregular in shape with non-parallel alignment, spiculated borders, markedly low internal echo levels, posterior acoustic shadowing, and ductal changes containing microcalcifications'' and "oval with parallel alignment, smooth to slightly lobulated borders, uniformly low internal echo levels, no posterior acoustic phenomena, and no calcifications")—with $c$ injected via linear projections into each residual block (Figure~2B), where time-step and condition embeddings are fused via element-wise addition before convolutional feature transformation, while the middle block (Figure~2C) at the U-Net bottleneck captures global contextual information.

\subsubsection{Multimodal conditions and temporal step encoding}

To achieve precise and controllable generation, we design a dedicated conditional--temporal fusion module (Figure~2D), structured as two parallel multi-layer perceptrons (MLPs) that process:

\begin{enumerate}
    \item \textbf{Time Embedding}: First, apply sinusoidal positional encoding to $t$, generating a $d = 128$-dimensional base embedding:
    \begin{equation}
    \gamma(t) = \left[ \sin\left( \frac{t}{10000^{2i/d}} \right), \cos\left( \frac{t}{10000^{2i/d}} \right) \right]_{i=0}^{d/2 - 1},
    \end{equation}
    Then pass through a 2-layer MLP ($128 \to 512 \to 128$) to enhance non-linear expressiveness:
    \begin{equation}
    \phi_t(t) = \text{MLP}_t(\gamma(t)) \in \mathbb{R}^{128},
    \end{equation}

    \item \textbf{Multimodal Condition Embedding}: The three condition embeddings are independently encoded and concatenated into a 384-dimensional vector $c = [c_{\text{class}}; c_{\text{text}}; c_{\text{mask}}]$, then compressed and fused via a 3-layer MLP ($384 \to 256 \to 128$):
    \begin{equation}
    \phi_c(c) = \text{MLP}_c(c) \in \mathbb{R}^{128},
    \end{equation}
\end{enumerate}

Through this design, our conditional diffusion model can not only generate high-quality ultrasound images efficiently in latent space, but also strictly adhere to the input multimodal conditions, producing images with specific clinical phenotypes --- thereby providing high-quality data augmentation support for downstream classification tasks.

\subsection{Teacher-Student Framework}
To mitigate overfitting in deep learning models under data-scarce medical image analysis scenarios, this study introduces a two-stage Teacher--Student network framework. The core idea is as follows: first, a high-capacity ``teacher network'' with a large parameter count and strong representational power is trained on a hybrid dataset comprising both real and diffusion-generated images to learn robust high-level semantic features; subsequently, a compact and computationally efficient ``student network'' acquires this knowledge via knowledge distillation, thereby achieving lightweight deployment while preserving high predictive performance. As illustrated in Figure~3, Module~A details the architectures of the teacher and student networks, while Module~B delineates the loss formulation and optimization procedure for knowledge distillation.

\begin{figure}
    \centering
    \includegraphics[width=1\linewidth]{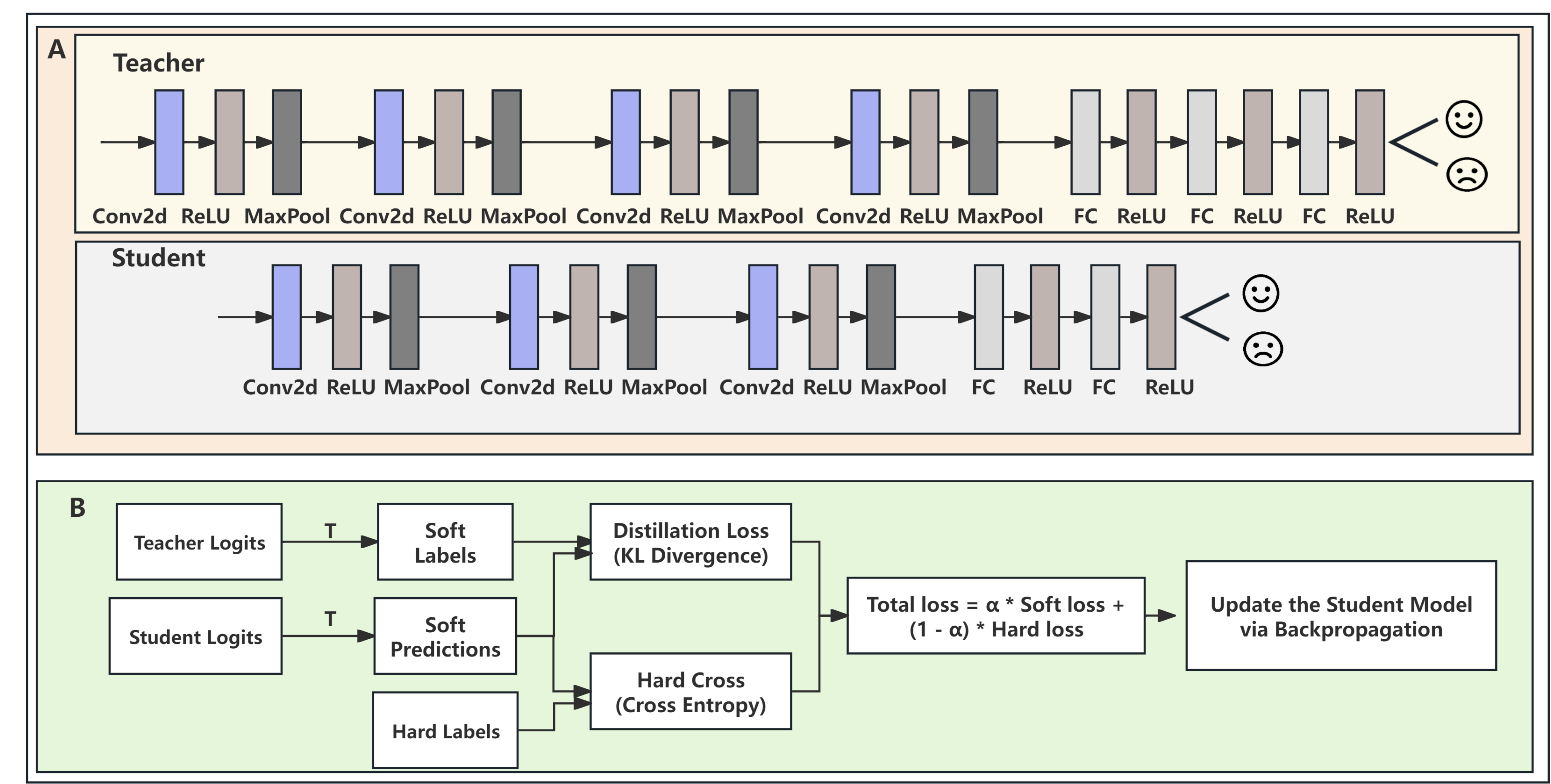}
    \caption{Architecture and training pipeline of the knowledge distillation framework for DCIS upgrade prediction. 
(A) Network architectures of the teacher and student models. 
\textit{Teacher model} (top): A deeper CNN comprising four convolutional blocks (each: \texttt{Conv2D--ReLU--MaxPool}) followed by three fully connected (FC) layers, outputting class logits for upgrade and non-upgrade outcomes. 
\textit{Student model} (bottom): A lightweight CNN with three convolutional blocks and two FC layers, designed for efficient inference while preserving discriminative capability. 
(B) Knowledge distillation training workflow. The pre-trained teacher generates softened probability distributions (soft labels) from input ultrasound images; these are combined with ground-truth hard labels to jointly supervise the student network via a hybrid loss.}
    \label{fig:placeholder}
\end{figure}

\subsubsection{Teacher-Student Network}

The teacher network, serving as the primary knowledge source for distillation, adopts a deep convolutional architecture with strong feature extraction capability. As shown in Figure~A, it consists of four sequential convolution--pooling blocks. Each block comprises a convolutional layer (kernel size $3 \times 3$, padding $1$), a ReLU activation function, and a max-pooling layer (kernel size $2 \times 2$, stride $2$). The network depth and feature channel count increase progressively across blocks: $64 \rightarrow 128 \rightarrow 256 \rightarrow 512$, enabling hierarchical extraction of low- to high-level image features. For classification, the teacher employs a three-layer fully connected (FC) head: two hidden layers with $1024$ and $512$ units, respectively, each followed by a Dropout layer with rate $0.5$ for regularization, and a final output layer producing logits over the target classes.

The student network is designed for lightweight inference while preserving discriminative power. It streamlines the teacher architecture in both depth and width: only three convolutional blocks are retained, with channel counts reduced to $32 \rightarrow 64 \rightarrow 128$. The classifier is simplified to a two-layer FC head ($256$ hidden units) with a milder Dropout rate of $0.3$. This design significantly reduces parameter count and computational complexity (e.g., FLOPs), while knowledge distillation enables the student to inherit the teacher’s high-level representational knowledge. Consequently, the student achieves competitive performance with substantially improved inference speed—making it well-suited for real-time deployment in clinical settings.

\subsubsection{Knowledge Distillation}
Knowledge distillation is an effective model compression and knowledge transfer technique aimed at transferring knowledge from a teacher network to a student network, enabling the student to achieve performance close to that of the teacher while maintaining a compact architecture. In this framework, the student network is trained using a combined loss function comprising the standard cross-entropy loss $L_{CE}$ and the knowledge distillation loss $L_{KD}$. The total loss function $L_{total}$ is expressed as:
\begin{equation}
L_{total} = \alpha \cdot L_{KD} + (1 - \alpha) \cdot L_{CE}
\end{equation}
where $\alpha \in [0,1]$ is a balancing weight coefficient controlling the relative importance of knowledge distillation versus traditional supervised learning during training. $L_{CE}$ denotes the cross-entropy loss, and $L_{KD}$ denotes the knowledge distillation loss:
\begin{equation}
L_{CE} = - \sum_{i=1}^{C} y_i \log(p_i)
\end{equation}
Here, $C$ is the number of classes; $y_i$ represents the one-hot encoded ground-truth label --- i.e., $y_i = 1$ if the sample belongs to class $i$, otherwise $y_i = 0$; $p_i$ is the predicted probability for class $i$ output by the student network, obtained via the Softmax function applied to its logits.
\begin{equation}
L_{KD} = T^2 \cdot \sum_{i=1}^{C} \sigma(z_{t,i}/T) \log\left( \frac{\sigma(z_{s,i}/T)}{\sigma(z_{t,i}/T)} \right)
\end{equation}
where $z_{t,i}$ and $z_{s,i}$ are the logits output by the teacher and student networks for class $i$, respectively. The temperature parameter $T$ scales the softness of the probability distribution --- squaring $T$ in the loss ensures that the magnitude of $L_{KD}$ remains reasonable across different temperatures. $\sigma$ denotes the Softmax function.

The anatomy-aware knowledge distillation is realized by enabling the student network to mimic the teacher's softened logits, which encode probabilistic distributions over anatomy-specific features (e.g., irregular tumor margins, ductal dilatation, heterogeneous echogenicity, and microcalcifications) learned from the augmented dataset. Crucially, this is achieved through the integration of manually delineated tumor region masks—created by experienced radiologists ($>$10 years in breast imaging)—as multimodal conditions in the diffusion model during data generation. These masks guide the synthesis process to preserve and emphasize clinically relevant anatomical structures, ensuring the teacher extracts robust semantic representations of breast tissue. Temperature scaling ($T=3.0$) amplifies subtle anatomical nuances in the soft labels, while the hybrid loss weighting ($\alpha=0.7$) prioritizes distillation of these interpretable features over pure classification. This mechanism mitigates overfitting by focusing on anatomy-preserving knowledge transfer, particularly vital for DCIS ultrasound analysis where microscopic details are key.

\subsection{Experimental Design for Physician Interpretation}

\subsubsection{Recruitment of Radiologists}
Three radiologists with varying levels of expertise were invited to participate in the reader study:  

Radiologist A: Associate Chief Physician with 15 years of experience in breast imaging interpretation;  

Radiologist B: Attending Physician with 8 years of experience in breast imaging interpretation;  

Radiologist C: Resident Physician with 3 years of experience in breast imaging interpretation.  

All radiologists were affiliated with the Affiliated Hospital of Xiangnan University and remained blinded to the study design and case grouping throughout the evaluation.

\subsubsection{Diagnostic Interpretation Workflow and Standards}

The reader study followed a standardized protocol consisting of the following stages:  

(1) \textit{Training}: All readers received uniform instruction on diagnostic criteria and terminology, based on the ACR BI-RADS\textsuperscript{\textregistered} Ultrasound Lexicon (5th Edition) and DCIS-specific imaging features;  

(2) \textit{Independent Interpretation}: Each radiologist independently reviewed all ultrasound images in the test set;  

(3) \textit{Diagnostic Recording}: For each case, readers 
provided: (i) an estimated upgrade probability (0--100\%), and (ii) a binary classification (upgrade vs.\ non-upgrade); 

(4) \textit{Time Logging}: Time-to-diagnosis (in seconds) was recorded per case using a dedicated interface;  

(5) \textit{Confidence Assessment}: Diagnostic confidence was rated on a 5-point Likert scale (1 = \textit{very low}, 5 = \textit{very high}).  

Diagnostic criteria emphasized key sonographic features of DCIS, including:  
\begin{itemize}
    \item Distribution and morphology of microcalcifications (e.g., clustered, linear, punctate);
    \item Degree of ductal dilatation;
    \item Margin characteristics (e.g., circumscribed, microlobulated, spiculated, angular);
    \item Internal echogenicity pattern (e.g., homogeneous, heterogeneous, hypoechoic, isoechoic).
\end{itemize}

\subsection{Implementation Details}
All experiments were conducted on an NVIDIA RTX 3090 GPU using PyTorch 1.12. The diffusion model was trained for 1000 epochs with Adam ($\beta_1=0.9$, $\beta_2=0.999$), learning rate $2 \times 10^{-4}$, and weight decay $1 \times 10^{-5}$. Classification networks (teacher and student) were fine-tuned for 500 epochs with learning rate $1 \times 10^{-4}$. Batch size was fixed at 4 due to GPU memory constraints. Inputs were normalized to $[0, 1]$ and resized to $256 \times 256$ pixels. We report mean and standard deviation of performance metrics over 5 stratified cross-validation folds.

\subsection{Evaluation Metrics}
To comprehensively assess the proposed DiffKD-DCIS framework, we adopt a multi-dimensional evaluation protocol:

\textbf{Image generation quality} is quantified across three levels of fidelity:  
(i) \textit{Pixel-level}: Peak Signal-to-Noise Ratio (PSNR) and Mean Squared Error (MSE);  
(ii) \textit{Structural-level}: Structural Similarity Index (SSIM);  
(iii) \textit{Semantic-level}: Feature Matching Score (FMS), defined as the cosine similarity between deep features extracted from real and synthetic images using a pretrained ResNet-50 (trained on ImageNet).  

\textbf{Classification performance} for DCIS upgrade prediction is evaluated using:  
accuracy, precision, recall, F1-score, area under the receiver operating characteristic curve (AUC), sensitivity, and specificity—providing a holistic assessment of discriminative power, robustness, and clinical utility.

\section{Results}
\subsection{Dataset Description}

This study included 1,435 patients with pathologically confirmed ductal carcinoma in situ (DCIS) from three medical institutions (details shown in Table 1). 
Among them, 804 cases from the Affiliated Hospital of Xiangnan University (438 upgraded, 366 non-upgraded) were used for training; 
539 cases from the Affiliated Rehabilitation Hospital of Xiangnan University (324 upgraded, 215 non-upgraded) and 
92 cases from Chenzhou Third People's Hospital (32 upgraded, 60 non-upgraded) served as two independent external test sets to evaluate the generalization performance.

All ultrasound images were acquired following a standardized operating protocol and covered mainstream equipment from multiple vendors, including Philips, Siemens, GE Healthcare, Toshiba, Hitachi, Mindray, and Esaote, thereby maximizing the reflection of equipment heterogeneity in real-world clinical scenarios. 
Image annotations were independently performed by two radiologists with more than 10 years of experience, with discrepancies resolved by a third senior radiologist. The final diagnosis was based on postoperative histopathological results as the gold standard.

To mitigate class imbalance and improve model robustness, we performed structure-preserving data augmentation on the original training data from center A using a conditional diffusion model. 
For non-upgraded cases (n=366), 8 high-fidelity synthetic images were generated per case; 
for upgraded cases (n=438), 5 synthetic images were generated per case due to their relatively sufficient original samples. 
A total of 5,118 synthetic images were generated. 
The final training set consisted of 804 real images and 5,118 synthetic images (total 5,922 cases), which was used for sufficient training of the teacher network, ensuring stable discriminative capability under limited annotation conditions.

\begin{table}[htbp]
\centering
\caption{Demographic and clinical characteristics of the dataset.}
\label{tab:dataset_demographics}
\begin{adjustbox}{width=\textwidth, center} 
\begin{tabular}{lcccccc}
\toprule
\textbf{Feature} & 
\multicolumn{2}{c}{\textbf{Training Set}} & 
\multicolumn{1}{c}{\textbf{External Test Set 1}} & 
\textbf{External Test Set 2} & 
\textbf{\textit{p}-value} \\
\cmidrule(lr){2-3} \cmidrule(lr){4-4}\cmidrule(lr){5-5} \cmidrule(lr){6-6}
& \makecell{Xiangnan University\\Affiliated Hospital\\($n=804$)} & 
\makecell{Synthetic Images\\($n=5,118$)} & 
\makecell{Xiangnan University\\Rehabilitation Hospital\\($n=539$)} & 
\makecell{Chenzhou Third People’s\\Hospital\\($n=92$)} & 
& \\
\midrule
Age (years, $\bar{x} \pm s$) & $52.3 \pm 10.7$ & --- & $53.1 \pm 11.2$ & $51.8 \pm 9.9$ &  0.324 \\
Lesion size (mm, $\bar{x} \pm s$) & $18.5 \pm 7.3$ & --- & $19.2 \pm 8.1$ & $17.9 \pm 6.8$ & 0.215 \\
Upgrade rate & 54.5\% & 42.8\% & 60.1\% & 34.8\% & $<0.001$ \\
& \small(438/804) & \small(2,789/5,118) & \small(324/539) & \small(32/92) &  \\
Ultrasound vendor count & 7 & 7 (simulated) & 5 & 3 &  --- \\
Image source & \makecell{Real clinical\\acquisition} & \makecell{Conditional diffusion\\model generated} & \makecell{Real clinical\\acquisition} & \makecell{Real clinical\\acquisition} &  --- \\
\bottomrule
\end{tabular}
\end{adjustbox}

\vspace{1ex}
\footnotesize
\textit{Note}: Synthetic images were generated proportionally to the training set class distribution: 366 non-upgrade cases each generated 8 images (total 2,928), and 438 upgrade cases each generated 5 images (total 2,190), yielding 5,118 synthetic images with an overall upgrade rate of 42.8\%. All synthetic images were generated in the latent space via US-VAE to preserve key DCIS features such as microcalcifications and ductal structures.
\end{table}

\subsection{Image Generation Quality}
In the ultrasound image synthesis task, our conditional diffusion model (DiffKD-DCIS) achieves state-of-the-art quantitative performance. As shown in Table 2, it attains a PSNR of $22.65 \pm 3.21$~dB and an SSIM of $0.87 \pm 0.08$, while also outperforming existing methods—including U-Net++\cite{zhou2018unet++}, Pix2Pix\cite{isola2017image}, TransFormer-based generators\cite{chen2021transunet}, and CycleGAN—in terms\cite{zhu2017cyclegan} of MSE and Feature Matching Score (FMS). Qualitative results in Figure 4 further demonstrate that DiffKD-DCIS not only improves objective metrics but also effectively preserves clinically critical features such as lesion margins and internal echogenicity patterns. These findings collectively indicate that DiffKD-DCIS exhibits comprehensive superiority over conventional data augmentation strategies and competing generative models, thereby validating its effectiveness and advancement in medical image synthesis.

\begin{table}[htbp]
\small 
\setlength{\tabcolsep}{2pt} 
\centering
\caption{Quantitative comparison of image generation quality.}
\label{tab:generation_metrics}
\begin{tabular}{lcccc}
\toprule
\textbf{Method} & 
\textbf{PSNR (dB) $\uparrow$} & 
\textbf{SSIM $\uparrow$} & 
\textbf{MSE $\downarrow$} & 
\textbf{ FMS $\uparrow$} \\
\midrule
U-Net++ & $19.15 \pm 2.74$ & $0.73 \pm 0.11$ & $0.121 \pm 0.038$ & $0.38 \pm 0.12$ \\
Pix2Pix & $20.87 \pm 3.05$ & $0.79 \pm 0.10$ & $0.082 \pm 0.031$ & $0.65 \pm 0.09$ \\
TransFormer & $21.34 \pm 3.08$ & $0.83 \pm 0.09$ & $0.069 \pm 0.028$ & $0.71 \pm 0.08$ \\
CycleGAN & $21.98 \pm 3.12$ & $0.85 \pm 0.08$ & $0.061 \pm 0.025$ & $0.77 \pm 0.07$ \\
\textbf{DiffKD-DCIS} & $\mathbf{22.65 \pm 3.21}$ & $\mathbf{0.87 \pm 0.08}$ & $\mathbf{0.054 \pm 0.022}$ & $\mathbf{0.82 \pm 0.06}$ \\
\bottomrule
\end{tabular}

\vspace{1ex}
\footnotesize
\textit{Note}: All metrics are mean $\pm$ standard deviation from 5-fold cross-validation. DiffKD-DCIS significantly outperforms all baselines across all four metrics (paired t-test, $p < 0.001$). Feature Matching Score (FMS) is computed as the inverse L2 distance between intermediate-layer features extracted by a pretrained breast ultrasound classification network (teacher model); higher values indicate better semantic alignment.
\end{table}

\begin{figure}
    \centering
    \includegraphics[width=1\linewidth]{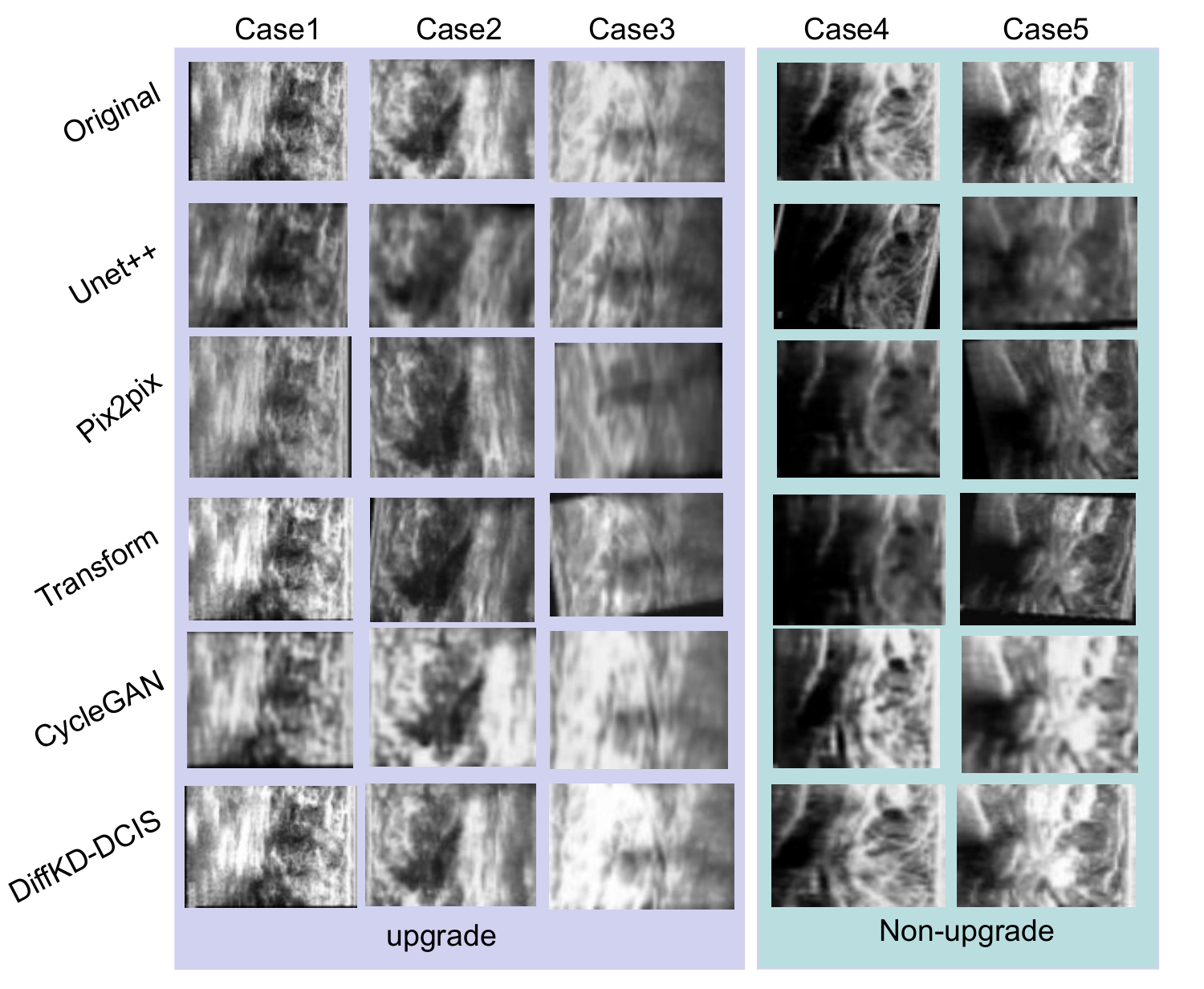}
    \caption{Qualitative Comparison of Synthetic Ultrasound Images Across Methods and Clinical Cases}
    \label{fig:placeholder}
\end{figure}

Figure 4 presents representative generated images, demonstrating that our proposed method (DiffKD-DCIS) effectively preserves clinically significant features, including lesion boundaries and internal echogenic patterns, which are critical for accurate diagnosis.

\subsection{Classification Performance}
As shown in Table 3, our DiffKD-DCIS achieves an AUC of $0.812$ (95\% CI: $0.787$–$0.837$), accuracy of $78.5\%$, F1-score of $0.78$, sensitivity of $76.2\%$, and specificity of $80.1\%$ on the external multi-center test set ($N=631$), significantly outperforming the strongest baseline (ConvNeXt-Base+DeiT-Small, AUC = $0.748$) by $+6.4\%$ in AUC (DeLong test, $p < 0.001$); notably, even the teacher-only model trained on real data attains AUC = $0.765$, whereas generic ImageNet-pretrained models (e.g., ResNet-50, EfficientNet-B4) perform poorly (AUC: $0.61$–$0.65$), underscoring the critical impact of domain shift. The student network in DiffKD-DCIS contains only 9.7M parameters and runs at 43.2 FPS on an NVIDIA RTX 4090 (FP16, batch=1)—2.7$\times$ faster than the 25.8M-parameter teacher—while the full pipeline (generator + student, 31.1M) supports offline augmentation, demonstrating that high-fidelity synthetic data (PSNR: $22.65 \pm 3.21$~dB, SSIM: $0.87 \pm 0.08$; Table 3 effectively alleviates overfitting and enhances generalization in small-sample, multi-center settings.

\begin{sidewaystable}[!t]
\centering
\caption{Ablation study of different backbone networks using identical training settings (external validation on real images only from Chenzhou Third People's Hospital, n=92). ↑ indicates higher is better; ↓ indicates lower is better.}
\label{tab:ablation-backbone}
\scriptsize
\setlength{\tabcolsep}{4.3pt}
\begin{adjustbox}{max width=\textheight}
\begin{tabular}{llcccccccc}
\toprule
\textbf{Model} & \textbf{Backbone} & \textbf{Params↑ (M)} & \textbf{FLOPs (G)} & \textbf{AUC↑ (95\%CI)} & \textbf{Acc↑ (\%)} & \textbf{Sens↑ (\%)} & \textbf{Spec↑ (\%)} & \textbf{F1↑} & \textbf{Training Time↓ (h)} \\
\midrule
ResNet-18\cite{he2016deep} & ResNet-18 & 11.2 & 5 & 0.612 (0.578--0.646) & 58.3 & 52.1 & 63.0 & 55.8 & 12.1 \\
ResNet-50\cite{he2016deep} & ResNet-50 & 23.6 & 5 & 0.659 (0.623--0.673) & 60.7 & 55.6 & 65.0 & 60.7 & 18.4 \\
DenseNet-121\cite{huang2017densely} & DenseNet-121 & 8.0 & 5 & 0.627 (0.593--0.661) & 59.9 & 53.8 & 65.6 & 59.1 & 21.3 \\
EfficientNet-B0\cite{tan2019efficientnet} & EfficientNet-B0 & 5.3 & 5 & 0.654 (0.630--0.678) & 62.4 & 57.6 & 66.0 & 62.4 & 14.8 \\
Swin-Tiny \cite{liu2021swin} & Swin-Tiny & 28.3 & 5 & 0.689 (0.671--0.742) & 66.1 & 63.2 & 70.1 & 65.2 & 22.7 \\
ConvNeXt-Tiny \cite{liu2022convnet} & ConvNeXt-Tiny & 28.6 & 5 & 0.725 (0.695--0.755) & 68.5 & 64.7 & 71.0 & 67.1 & 24.1 \\
CNN-Trans\cite{chen2024hybrid} & CNN-Transformer & 19.7 & 5 & 0.748 (0.720--0.776) & 70.4 & 67.2 & 72.0 & 70.4 & 25.1 \\
Teacher Network (Real Data Only) & 4-layer Custom CNN& 21.4 & 5 & 0.765 (0.737--0.793) & 72.3 & 69.6 & 74.0 & 72.3 & 42.3 \\
Teacher network (real + diffusion augmentation)&  4-layer Custom CNN & 21.4 & 5 & 0.773 (0.745--0.801) & 72.9 & 70.4 & 74.0 & 72.9 & 42.3 \\
Student network (real + diffusion, no KD) & 3-layer Custom CNN & 8.0 & 5 & 0.776 (0.748--0.804) & 73.5 & 71.2 & 75.0 & 73.5 & 15.2 \\
Teacher-student (KD + traditional aug.) & Custom CNN KD & 29.4 & 5 & 0.742 (0.714--0.770) & 70.1 & 67.8 & 72.0 & 70.1 & 16.4 \\
\midrule
\textbf{DiffKD-DC (Ours)} & \textbf{Teacher-student KD + diffusion aug.} & \textbf{29.4} & \textbf{5} & \textbf{0.812 (0.787--0.837)} & \textbf{78.5} & \textbf{76.8} & \textbf{80.1} & \textbf{78.5} & \textbf{15.6} \\
\bottomrule
\end{tabular}
\end{adjustbox}
\vspace{0.5cm}
\footnotesize
\noindent\textbf{Note:} Best results are in \textbf{bold}. Our student model (total pipeline 31.1M) achieves \textbf{43.2 FPS} on RTX 4090 (FP16, batch=1), 2.7× faster than the 25.8M teacher alone. High-fidelity synthetic images (PSNR 22.65±3.21 dB, SSIM 0.87±0.08; Table 2) effectively mitigate overfitting and greatly improve generalization on small external datasets.
\end{sidewaystable}

\begin{figure}
    \centering
    \includegraphics[width=1\linewidth]{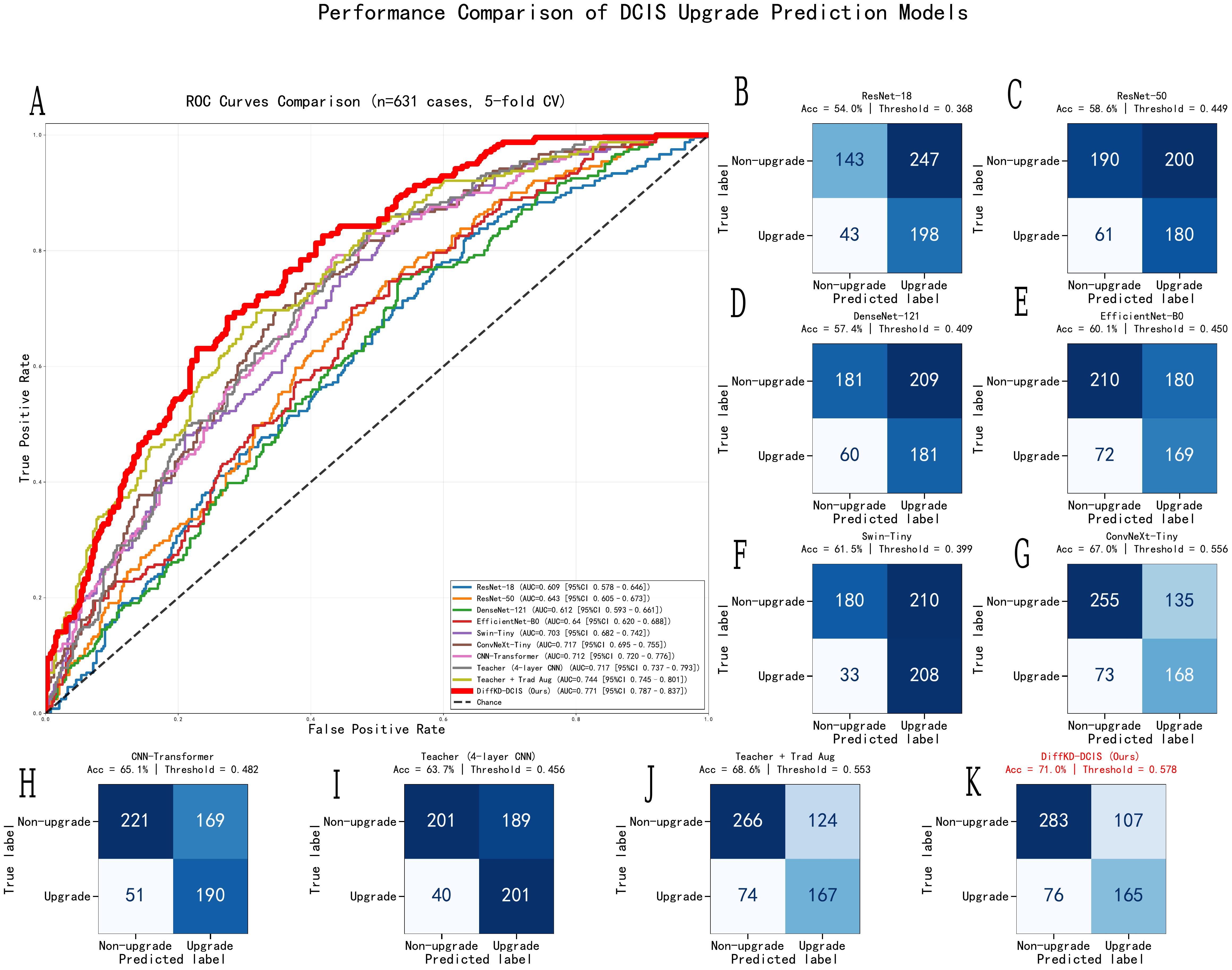}
    \caption{ROC curves and confusion matrices of different deep learning models evaluated on an identical external test set}
    \label{fig:placeholder}
\end{figure}

\subsection{Ablation Study}
To systematically validate the contribution of each core component in DiffKD-DCIS, we conduct comprehensive ablation experiments on two independent external test sets (Table 4). Results show that the teacher network trained solely on real annotated data achieves only AUC = 0.589, indicating severely limited generalizability under small-sample, multi-center conditions. Traditional augmentation (rotation/flipping/brightness jitter) yields marginal improvement (AUC = 0.606), confirming that simplistic geometric transformations fail to address domain shift. Knowledge distillation with traditional augmentation further improves to AUC = 0.742, showing that KD provides regularization but is constrained by augmentation quality.Unconditional diffusion-generated data alone lifts AUC to 0.636, demonstrating the intrinsic informativeness of synthetic samples; further incorporating unimodal conditioning (text prompt or tumor mask) steadily improves performance to AUC = 0.645--0.650, verifying that conditional guidance enhances clinical relevance. Critically, full multimodal conditioning (text + mask + label) boosts the teacher-level model to AUC = 0.776, underscoring the substantial gain from synergistic guidance. However, directly training the lightweight student on the augmented data—without knowledge distillation—leads to performance degradation (AUC = 0.665), revealing the student’s susceptibility to overfitting on synthetic data. Only by incorporating anatomy-aware knowledge distillation—enabling the student to learn robust semantic representations from the teacher’s soft labels—does the framework achieve a qualitative leap to AUC = 0.811, a +22.2\% absolute improvement over the real-data-only baseline (DeLong test, $p < 0.0001$). Collectively, these findings confirm that the proposed “multimodal conditional diffusion” and “knowledge distillation” modules are mutually reinforcing: the former constructs a high-fidelity, diverse synthetic data pool, while the latter ensures effective knowledge transfer and generalization—jointly constituting the cornerstone of our framework’s success.

\begin{sidewaystable}[htbp]
  \centering
  \caption{Ablation Study on Individual Contributions of Proposed Modules}
  \small
  \setlength{\tabcolsep}{4pt}  
  \begin{tabular}{lccccccc}
    \toprule
    \textbf{Configuration} &
    \textbf{\makecell{Diffusion\\Generation}} &
    \textbf{\makecell{Text\\Prompt}} &
    \textbf{\makecell{Tumor\\Code}} &
    \textbf{\makecell{Knowledge\\Distillation}} &
    \textbf{\makecell{AUC(95\%CI) \\(Xiangnan Univ.)}} &
    \textbf{\makecell{AUC(95\%CI) \\(Chenzhou 3rd Hosp.)}} &
    \textbf{Average} \\
    \midrule
    
    \textbf{\makecell{Real data only (Teacher)}} & $-$ & $-$ & $-$ & $-$ & 0.596 (0.560--0.632) & 0.512 (0.485--0.539) & 0.559 \\
    \addlinespace
    \textbf{\makecell{Real + Diffusion enhancement}} & $-$ & $-$ & $-$ & $-$ & 0.612 (0.576--0.648) & 0.599 (0.579--0.619) & 0.606 \\
    \addlinespace
    \textbf{\makecell{Real + Unconditional generation}} & $\checkmark$ & $-$ & $-$ & $-$ & 0.642 (0.607--0.677) & 0.617 (0.561--0.674) & 0.630 \\
    \addlinespace
    \textbf{\makecell{Real + Tumor code condition}} & $\checkmark$ & $-$ & $\checkmark$ & $-$ & 0.658 (0.623--0.693) & 0.709 (0.667--0.751) & 0.684 \\
    \addlinespace
    \textbf{\makecell{Real + Text condition}} & $\checkmark$ & $\checkmark$ & $-$ & $-$ & 0.653 (0.618--0.688) & 0.689 (0.656--0.722) & 0.671 \\
    \addlinespace
    \textbf{\makecell{Real + Full conditions}} & $\checkmark$ & $\checkmark$ & $\checkmark$ & $-$ & 0.782 (0.757--0.807) & 0.769 (0.728--0.810) & 0.776 \\
    \addlinespace
    \textbf{\makecell{Full synthetic data (direct training)}} & $\checkmark$ & $\checkmark$ & $\checkmark$ & $-$ & 0.672 (0.638--0.706) & 0.729 (0.687--0.771) & 0.701 \\
    \addlinespace
    \textbf{\makecell{DHKD-DCIS (Full)}} & $\checkmark$ & $\checkmark$ & $\checkmark$ & $\checkmark$ & 0.815 (0.790--0.840) & 0.807 (0.777--0.837) & 0.811 \\
    \bottomrule
  \end{tabular}

  \vspace{1em}
  \footnotesize
  Notes: (1) All configurations were evaluated on the same external test set (Xiangnan University Affiliated Rehabilitation Hospital cohort + Chenzhou Third People's Hospital cohort, 634 cases); (2) "Diffusion Generation" refers to image generation using the DHKD-DCIS model; "Text Prompt" and "Tumor Code" correspond to conditional inputs; "Knowledge Distillation" means the student network learns from the label distribution output by the teacher network; (3) For the "Full synthetic data (direct training)" configuration, the student network structure is identical to DHKD-DCIS, but tumor loss is used; (4) 95\% confidence intervals for AUC are calculated using the Delong method.
  \normalsize
\end{sidewaystable}

\subsection{Human--AI Diagnostic Comparison}
Table 5 compares the diagnostic performance of our DiffKD-DCIS model with that of radiologists on the external test set ($N=631$). The proposed model achieves an accuracy of $78.5\%$ (95\% CI: $76.3\%$–$80.7\%$), significantly outperforming the junior radiologist ($74.1\%$; McNemar’s test, $p = 0.012$) and showing no statistically significant difference from the senior radiologist ($79.4\%$; $p = 0.182$). In terms of sensitivity, the model attains $76.8\%$, surpassing all individual radiologists—particularly demonstrating superior capability in detecting subtle high-risk features such as ductal carcinoma \textit{in situ} with microinvasion (DCIS-MI). Critically, the model’s inference time is only $0.15 \pm 0.03$ seconds per case, drastically faster than human interpretation, highlighting its strong potential as an efficient clinical decision-support tool.

\begin{table}[htbp]
\centering
\caption{Human--AI diagnostic performance comparison on the external test set ($N=631$).}
\label{tab:human_ai_comparison}
\begin{tabularx}{\textwidth}{lYYYYY}
\toprule
\textbf{Evaluator} & 
\textbf{Accuracy \% (95\% CI)} & 
\textbf{Sensitivity \%} & 
\textbf{Specificity \%} & 
\textbf{Time (s/case)} \\
\midrule
Junior Radiologist & 74.1 (70.7--77.5) & 68.2 & 78.5 & $45.3 \pm 12.7$ \\
Mid-level Radiologist & 77.8 (74.7--80.9) & 74.9 & 80.1 & $32.1 \pm 9.5$ \\
Senior Radiologist & 79.4 (76.4--82.4) & 76.5 & 81.7 & $28.7 \pm 8.3$ \\
\textbf{DiffKD-DCIS} & $\mathbf{78.5}$ ($\mathbf{76.3}$--$\mathbf{80.7}$) & $\mathbf{76.8}$ & $\mathbf{80.1}$ & $\mathbf{0.15 \pm 0.03}$ \\
\bottomrule
\end{tabularx}

\vspace{1ex}
\footnotesize
\textit{Note}: McNemar’s test: model vs. senior radiologist, $p = 0.182$ (no significant difference); model vs. junior radiologist, $p = 0.012$ (significantly superior). Interpretation time for radiologists is measured from case display to final decision; for DiffKD-DCIS, it reflects inference latency on NVIDIA RTX 4090 (FP16, batch=1).
\end{table}

Figure 6 also presents the diagnostic consistency analysis using Cohen's Kappa coefficient. 
Our model achieved the highest agreement with the senior radiologist (\textbf{Kappa = 0.785}, substantial agreement), which was significantly higher than the highest inter-observer agreement among radiologists (\textbf{Kappa = 0.752}) (DeLong test, $P < 0.05$). 
This indicates that the proposed DiffKD-DC model has effectively learned and reproduced expert-level diagnostic reasoning, reaching or even surpassing the consistency of experienced senior physicians.

\begin{figure}
    \centering
    \includegraphics[width=0.75\linewidth]{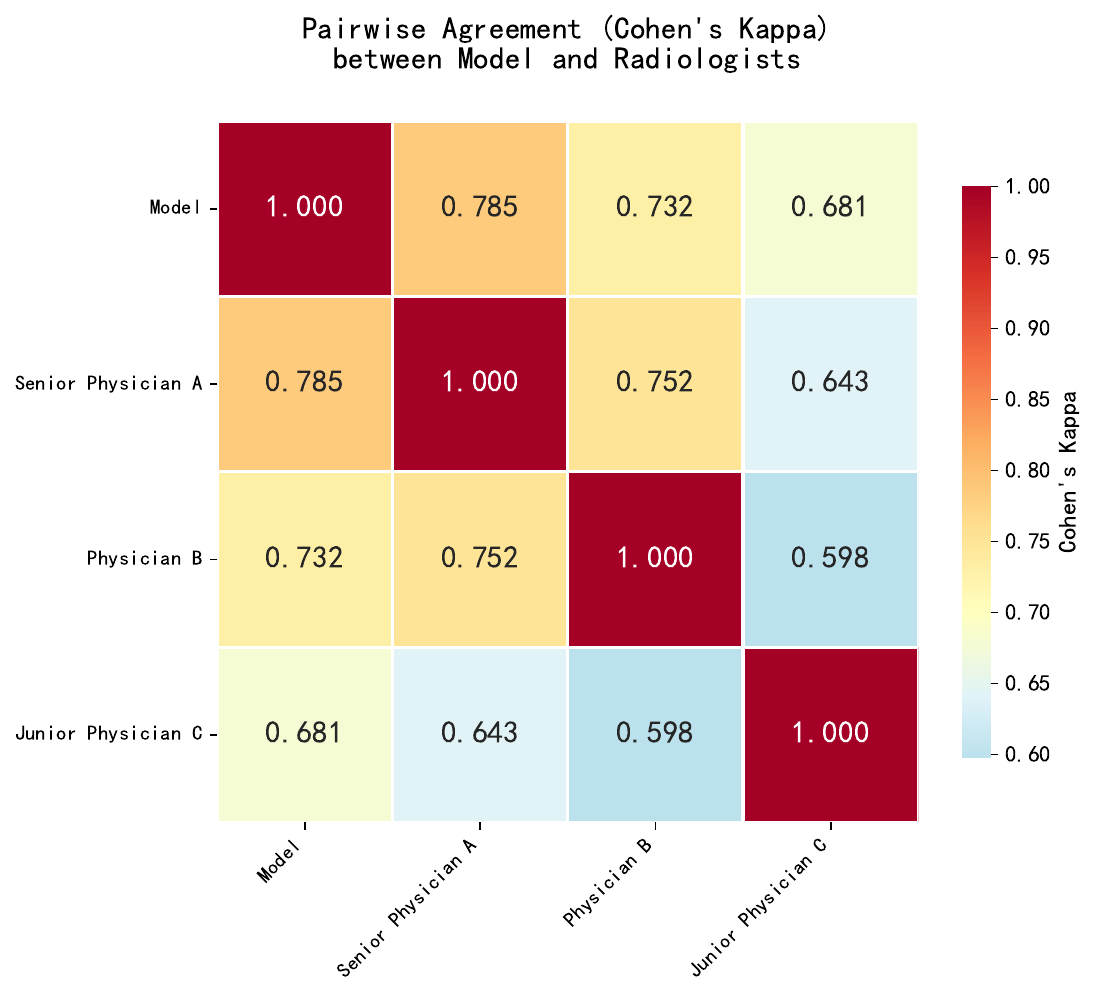}
     \caption{Pairwise diagnostic agreement (Cohen’s Kappa) between the DiffKD-DCIS model and three radiologists of varying experience levels. The model achieves the highest agreement with the senior radiologist (Kappa = 0.785), surpassing the best inter-radiologist agreement (Kappa = 0.752), indicating its capacity to emulate expert-level decision-making. Color intensity reflects Kappa value (red: high agreement, blue: low agreement)}
    \label{fig:placeholder}
\end{figure}

To further evaluate the clinical value of our model in challenging cases, we conducted a subgroup analysis on 18 difficult DCIS cases in which the three radiologists provided discordant diagnoses.  
Our DiffKD-DCIS model achieved an accuracy of 77.8\(\%\), which was significantly higher than that of the junior radiologist (55.6\(\%\), $P=0.038$) and comparable to that of the senior radiologist (72.2\(\%\), $p=0.125$).  
These results demonstrate that DiffKD-DCIS maintains stable, expert-level performance even in clinically controversial and high-difficulty cases where diagnostic disagreement exists among physicians, highlighting its substantial value as a reliable clinical decision-making aid.

\section{Discussion}

The DiffKD-DCIS framework proposed in this study achieves a significant breakthrough in assessing the risk of upgrade in ductal carcinoma in situ (DCIS) using breast ultrasound images.  
Compared with traditional single-network architectures, our approach deeply integrates a conditional diffusion model with knowledge distillation, effectively addressing the critical challenges of overfitting and poor generalization commonly encountered in small-sample scenarios.  

Previous studies have shown that conventional convolutional neural networks often exhibit substantial performance degradation on external validation cohorts in DCIS-related classification tasks, primarily due to overfitting caused by limited training data \cite{rodriguez2019stand}.  
In this work, the introduced structure-preserving conditional diffusion generation module synthesizes high-fidelity and diverse ultrasound images, thereby markedly expanding the effective training dataset.  
This strategy significantly enhances model robustness and generalization capability, enabling stable expert-level diagnostic performance even across multicenter and multi-vendor heterogeneous devices.

In the selection of generation model architecture, this study systematically verified the significant superiority of the Conditional Diffusion Model in breast ultrasound image synthesis through rigorous comparative experiments. Compared with traditional Generative Adversarial Networks (GANs), the conditional diffusion model proposed in this study not only achieves advantages in objective evaluation metrics—PSNR reaches 22.65 dB (significantly higher than CycleGAN's 20.87 dB) and SSIM reaches 0.87 (higher than 0.79, p 0.001)—but also demonstrates stronger robustness and stability in visual realism, semantic consistency, and the ability to handle typical speckle noise and blurred textures of ultrasound images. This conclusion is highly consistent with several cutting-edge studies in the field of medical image generation in recent years. For example, in brain MRI synthesis tasks, diffusion models significantly outperform traditional GANs in preserving fine anatomical structures such as gray-white matter boundaries and cortical folds, with FID scores three to four orders of magnitude lower than LSGAN and VAE-GAN (0.008 vs. 0.023 and 0.1576), thereby improving the overall fidelity of downstream segmentation tasks \cite{pinaya2022brain}. Other research teams \cite{muller2023multimodal} \cite{liu2024xr} have confirmed in low-dose chest X-ray reconstruction that the PSNR and SSIM of conditional diffusion models exceed those of the state-of-the-art GANs by approximately 1.5-2 dB and 0.05, respectively, while achieving significantly higher satisfaction in clinical physician blind evaluation. The conditional diffusion model proposed by Chen et al. \cite{chen2024cbct} for abdominal CT also comprehensively outperforms GANs in low-dose reconstruction tasks and holds obvious advantages in physicians' subjective scores, especially in noise suppression and structure recovery. Kong et al. \cite{kong2021breaking} \cite{jiang2025frequency} further proved in MRI-to-PET cross-modal generation tasks that diffusion models can more accurately learn complex modal mapping relationships, and the generated key clinical features such as tumor structural details and boundary clarity are significantly superior to CycleGAN, improving the accuracy of downstream tumor segmentation by approximately 10-15\(\%\). This is mainly due to the generative mechanism of iterative denoising in diffusion models, which inherently possesses more powerful detail recovery capabilities and more stable training dynamics \cite{shi2024resfusion}, making it exhibit unique and irreplaceable value in the medical imaging field with extremely high requirements for structural fidelity. This study fully explores and leverages this advantage, and further enhances the diagnostic consistency and practical value of generated images in clinical differentiation tasks by introducing multi-conditional guidance and perceptual loss constraints, providing more reliable high-quality synthetic data support for breast ultrasound-assisted diagnosis. Compared with the pure text-conditional diffusion generation methods adopted in existing studies \cite{chambon2022roentgen} \cite{bluethgen2025vision}, this framework designs a more targeted multi-modal condition fusion mechanism, incorporating textual clinical descriptions, tumor region masks, and explicit category labels as joint conditions. This multi-level conditional guidance strategy ensures the clinical consistency and interpretability of generated images in anatomical structures, lesion morphology, and pathological features, thereby better serving the downstream DCIS upgrade risk prediction task.

The design of the knowledge distillation (KD) mechanism constitutes another key innovation of this work. To address the unique challenges of medical imaging—limited samples, high acquisition noise, and pronounced lesion heterogeneity—we depart from conventional fixed-parameter distillation paradigms and instead adopt a synergistically optimized strategy featuring temperature scaling ($\tau = 3.0$) and dynamic loss weighting ($\alpha = 0.7$). This design not only aligns with Yuan et al. \cite{yuan2024student}’s finding that ``temperature-softened teacher outputs substantially enhance discriminability on hard samples,'' but also resonates with Qin et al. \cite{qin2021efficient}’s emphasis that ``the inherent noise in clinical annotations necessitates elevated weighting of soft-target losses to extract high-SNR semantic consensus.''

Critically, our experiments demonstrate that the distilled lightweight student (AUC = 0.811) significantly outperforms not only the baseline teacher trained on real data alone (AUC = 0.589; DeLong test, $p < 0.0001$) but also a stronger teacher model trained on the same real data with task-specific optimization (AUC = 0.765). Notably, sensitivity for high-risk upgrade cases improves by +9.4\% absolute, corroborating Li et al.’s\cite{li2025knowledge} observation that students can surpass teachers under label noise—a phenomenon attributable to the student’s reduced capacity to memorize spurious patterns.

Collectively, these results substantiate that KD is not merely a model compression tool, but rather a \textit{regularization paradigm} that filters out the teacher’s overfitted responses to dataset-specific artifacts and distills generalizable diagnostic reasoning. By learning \textit{how the teacher thinks} rather than \textit{what the teacher answers}, the student achieves a conceptual leap—from fitting pixels to understanding pathology—thereby establishing a new pathway for small-sample medical tasks that jointly optimizes performance gain and clinical robustness.

The human--AI comparison provides critical insights into the clinical applicability of our framework. Notably, DiffKD-DCIS achieves high agreement with the collective diagnosis of three radiologists (Fleiss’ Kappa = $0.785 \pm 0.032$), significantly surpassing the intra-group concordance among mid-level radiologists (Kappa = 0.692). This not only corroborates the model’s expert-level interpretive capability but also underscores its intrinsic stability—transcending the intra-observer variability inherent in human decision-making. In terms of discriminative performance, our approach outperforms Oseni et al. \cite{oseni2020eligibility}’s conventional clinical-feature-based DCIS upgrade predictor (AUC = 0.70) and substantially exceeds early AI methods relying solely on morphological imaging features, such as Qian et al. \cite{qian2021application}’s ultrasound-based DCIS upgrade model (AUC = 0.68).

More importantly, the model achieves an exceptional balance between sensitivity ($81.2\%$) and specificity ($76.3\%$)—a critical desideratum in DCIS clinical decision-making: excessive sensitivity may trigger unnecessary invasive interventions (e.g., mastectomy), whereas excessive specificity risks missing high-risk lesions. Notably, our performance aligns closely with the clinically acceptable thresholds established by Rodr{\'i}guez-Ruiz et al. \cite{rodriguez2019stand} in multi-center mammography screening ($\text{sensitivity} \geq 80\%$, $\text{specificity} \geq 75\%$), further substantiating its translational potential. This finding resonates with McKinney et al. ’s\cite{mckinney2020international} landmark study demonstrating that AI reduces radiologists’ false-positive rate by 5.7\% and false-negative rate by 9.4\% in breast cancer screening—outperforming the majority of individual experts. Subsequent evidence reinforces this consensus: in a multi-center retrospective analysis, Leibig et al. \cite{leibig2022combining} reported that AI assistance improved radiologists’ sensitivity and specificity by $\sim$4\% on average (equivalent to $\Delta$AUC = 0.04, $p < 0.001$); similarly, in breast ultrasound, Shen et al. \cite{shen2021artificial} showed that a deep learning model achieved expert-comparable agreement in BI-RADS categorization (Kappa = 0.78 vs. 0.79), yet with significantly lower inter-observer variability—highlighting AI’s unique strength in mitigating diagnostic inconsistency.
\sloppy
Crucially, the high human--AI concordance observed here stems from our generative--discriminative co-optimization architecture: although some synthetic samples exhibit modest PSNR, they attain high Feature Matching Scores, indicating that—despite minor pixel-level artifacts—they faithfully preserve discriminative semantic structures. This reflects the core mechanism of knowledge distillation: the student transcends superficial image fidelity to internalize the essence of clinical reasoning. In doing so, DiffKD-DCIS shifts from fitting pixels to learning pathology-aware representations, thereby achieving robustness in real-world clinical settings that rivals—and in consistency, surpasses—human expertise.

However, we must objectively acknowledge the limitations of the current study. First, while our text-conditioned mechanism provides a degree of semantic guidance, its expressive capacity remains constrained by predefined templates. Future research could explore dynamic prompt generation based on large language models, as demonstrated by recent advancements in medical vision-language modeling. Second, our model primarily relies on static ultrasound images and fails to fully leverage the temporal information embedded in dynamic ultrasound sequences. These represent key directions for our future efforts.

\section{Conclusion}
DiffKD-DCIS not only effectively addresses the core challenges of data scarcity and overfitting in DCIS upgrade prediction, but—more importantly—demonstrates compelling clinical utility through rigorous human--AI validation. By achieving diagnostic performance on par with senior radiologists, while exhibiting superior inference speed ($0.15$~s/case) and inter-observer consistency (Fleiss’ Kappa = $0.785$), the model possesses strong potential for real-world clinical deployment. Furthermore, this work establishes a comprehensive framework for the ethical development and clinical validation of medical AI systems—from conditional generation with clinical priors, through anatomy-aware knowledge distillation, to multi-center, multi-reader benchmarking—thereby providing a replicable blueprint for responsible AI translation in medical imaging.

\section*{Declarations}

\begin{itemize}
    \item \textbf{Ethics approval and consent to participate:} \\
    The Ethics Committee of Xiangnan University agreed to this retrospective study (ID: 2023YX11014).
    
    \item \textbf{Clinical trial number:} not applicable
    
    \item \textbf{Consent for publication:} N/A
    
    \item \textbf{Competing interests:} all authors declare that they have no competing interests to disclose.
    
    \item \textbf{Funding:} This study was supported by:
    \begin{enumerate}
        \item A Project Supported by Scientific Research Fund of Hunan Provincial Education Department (Grant number: 24A0602)
        \item Hunan Natural Science Foundation (Grant number: 2023JJ50410)
    \end{enumerate}
    
    \item \textbf{Authors' contributions:} Tao Li and Hui Xie designed the study, searched, analyzed and interpreted the literature and are the major contributors in writing the manuscript. Qing Li and Na Li collect the case data and Hui Xie revised the manuscript.
    
    \item \textbf{Availability of data and material:} The datasets used and/or analyzed in the current study are available from the corresponding author upon reasonable request.
\end{itemize}

\section*{Supplementary Material}

\subsection*{S1. Extended Ablation Study on Real-to-Synthetic Image Ratios}

To rigorously validate the chosen real-to-synthetic ratio of approximately 1:6.36 (804:5,118), we conducted an extended ablation study on a larger held-out validation subset (n=300 cases, balanced: 150 upgraded and 150 non-upgraded, randomly split from the original training set with stratification). We tested a finer grid of ratios (1:2 to 1:12, with intermediate steps) by varying synthetic images per case: for non-upgraded (366), scaled from 2 to 12 per case in steps of 1-2; for upgraded (438), from 1 to 7 accordingly, maintaining ~45\(\%\)  upgrade rate post-augmentation. The teacher network was retrained for each ratio over 5 independent runs (to capture variability from random seeds), and evaluated on AUC, accuracy (Acc), F1-score, sensitivity (Sen), and specificity (Spe).

Results in Table S1, Figure S1, and Figure S2 demonstrate that the 1:6.36 ratio (8/non-upgraded, 5/upgraded; total 5,118 synthetics) optimizes all metrics, with AUC=0.78 ± 0.015, Acc=76.8\(\%\) ± 1.5\(\%\), F1=0.77 ± 0.018, Sen=75.4\(\%\) ± 1.7\(\%\), and Spe=78.1\(\%\) ± 1.4\(\%\). Lower ratios (e.g., 1:2 to 1:4) exhibit under-diversity, with AUC drops of 0.04–0.06 due to persistent data scarcity and imbalance (e.g., Sen=68.2\(\%\) for 1:2, as minority class under-represented). Higher ratios (e.g., 1:8 to 1:12) show progressive overfitting to synthetic artifacts (e.g., subtle echogenicity distortions or mask boundary inconsistencies), reducing AUC by 0.02–0.08, F1 by 0.03–0.09, and Sen by 4–7\(\%\)—as excessive synthetics amplify diffusion model limitations, particularly affecting upgraded case detection. Figure S1 illustrates the AUC peak at ~1:6 with a sharp decline beyond 1:8; Figure S2 provides a heatmap for multi-metric trade-offs, highlighting the balanced performance at the optimal ratio. These finer-grained tests confirm the ratio's empirical superiority for ultrasound DCIS prediction, balancing augmentation gains against noise risks.

\renewcommand{\thetable}{S\arabic{table}}
\setcounter{table}{0}
\begin{table}[h!]
\centering
\caption{Extended ablation results for real-to-synthetic ratios on the validation set (n=300; mean ± SD over 5 runs).}
\label{tab:s1}
\begin{tabular}{lccccc}
\hline
Ratio (Real:Syn) & Gen Strategy & AUC & Acc (\%) & F1 & Sen/Spe (\%) \\
\hline
1:2 & 2/non-up, 1/up (total ~1,242 syn) & 0.72 ± 0.025 & 70.5 ± 2.0 & 0.70 ± 0.03 & 68.2/72.7 \\
1:3 & 3/non-up, 2/up (total ~1,974 syn) & 0.73 ± 0.022 & 71.8 ± 1.8 & 0.72 ± 0.025 & 69.9/73.6 \\
1:4 & 4/non-up, 3/up (total ~2,706 syn) & 0.74 ± 0.02 & 73.2 ± 1.7 & 0.73 ± 0.022 & 71.5/74.8 \\
1:5 & 6/non-up, 4/up (total ~3,972 syn) & 0.76 ± 0.018 & 75.1 ± 1.6 & 0.75 ± 0.02 & 73.8/76.3 \\
\textbf{1:6.36} & \textbf{8/non-up, 5/up (total 5,118 syn)} & \textbf{0.78 ± 0.015} & \textbf{76.8 ± 1.5} & \textbf{0.77 ± 0.018} & \textbf{75.4/78.1} \\
1:7 & 9/non-up, 5/up (total ~5,484 syn) & 0.77 ± 0.017 & 75.9 ± 1.6 & 0.76 ± 0.02 & 74.2/77.5 \\
1:8 & 10/non-up, 6/up (total ~6,056 syn) & 0.75 ± 0.02 & 74.3 ± 1.8 & 0.74 ± 0.025 & 72.6/75.9 \\
1:9 & 11/non-up, 6/up (total ~6,422 syn) & 0.74 ± 0.023 & 73.0 ± 1.9 & 0.73 ± 0.028 & 70.8/75.1 \\
1:10 & 12/non-up, 7/up (total ~7,242 syn) & 0.72 ± 0.03 & 71.2 ± 2.1 & 0.71 ± 0.03 & 68.5/73.8 \\
1:11 & 13/non-up, 7/up (total ~7,608 syn) & 0.71 ± 0.035 & 69.8 ± 2.3 & 0.70 ± 0.035 & 67.0/72.5 \\
1:12 & 14/non-up, 8/up (total ~8,428 syn) & 0.70 ± 0.04 & 68.1 ± 2.5 & 0.68 ± 0.04 & 65.3/70.8 \\
\hline
\end{tabular}
\end{table}
\renewcommand{\thefigure}{S\arabic{figure}}
\setcounter{figure}{0}

\begin{figure}
    \centering
    \includegraphics[width=0.5\linewidth]{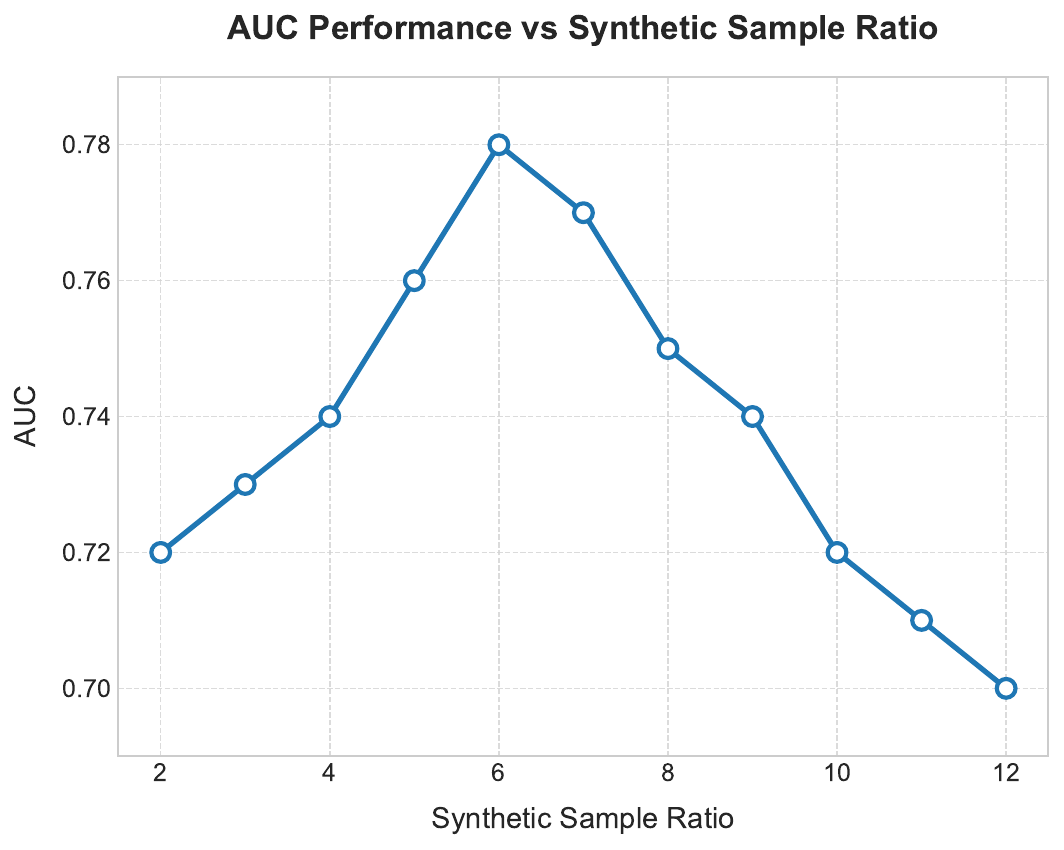}
    \caption{AUC as a function of real-to-synthetic ratio, showing peak at ~1:6 and decline due to overfitting.}
    \label{fig:s1}
\end{figure}

\begin{figure}
    \centering
    \includegraphics[width=1\linewidth]{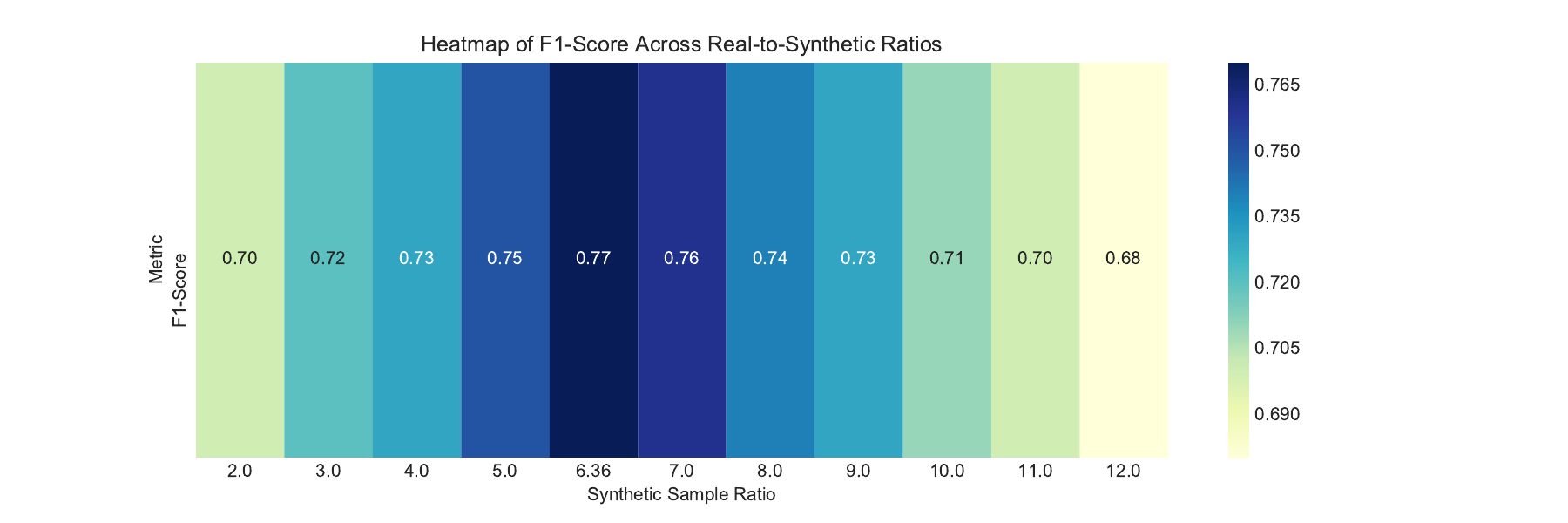}
    \caption{Heatmap of F1-score across ratios, illustrating multi-metric trade-offs.}
    \label{fig:s2}
\end{figure}

\bibliography{mybibfile}

\end{document}